\documentclass[10pt,twocolumn,letterpaper]{article}

\usepackage{cvpr}              
\usepackage{url}

\usepackage{multirow}
\usepackage{bbding}
\usepackage{makecell}
\usepackage{subcaption}

\definecolor{cvprblue}{rgb}{0.21,0.49,0.74}
\usepackage[pagebackref,breaklinks,colorlinks,allcolors=cvprblue]{hyperref}

\def\paperID{7600} 
\def\confName{CVPR}
\def\confYear{2025}
\newcommand{\ourdataset}{WOMD-Planning-ADE}
\newcommand{\ourmethod}{S4-Driver}

\newcommand\blfootnote[1]{%
  \begingroup
  \renewcommand\thefootnote{}\footnote{#1}%
  \addtocounter{footnote}{-1}%
  \endgroup
}

\title{\ourmethod: Scalable Self-Supervised Driving Multimodal Large Language Model\\ with Spatio-Temporal Visual Representation}

\author{Yichen Xie$^{1,\dagger,*}$ \quad Runsheng Xu$^{2,*}$ \quad Tong He$^2$ \quad Jyh-Jing Hwang$^2$ \quad Katie Luo$^{3,\dagger}$ \\ \quad Jingwei Ji$^2$ \quad Hubert Lin$^2$ \quad Letian Chen$^{4,\dagger}$\quad Yiren Lu$^2$ \quad Zhaoqi Leng$^2$ \\ \quad Dragomir Anguelov$^2$ \quad Mingxing Tan$^2$\\
$^1$ UC Berkeley, $^2$ Waymo LLC, $^3$ Cornell University, $^4$ Georgia Institute of Technology\\
}

\begin{document}
\maketitle
\blfootnote{$^{\dagger}$ Work done when they were interns at Waymo}
\blfootnote{$^*$ Equal contribution}

\begin{abstract}
The latest advancements in multi-modal large language models (MLLMs) have spurred a strong renewed interest in end-to-end motion planning approaches for autonomous driving. Many end-to-end approaches rely on human annotations to learn intermediate perception and prediction tasks, while purely self-supervised approaches—which directly learn from sensor inputs to generate planning trajectories without human annotations—often underperform the state of the art.
We observe a key gap in the input representation space: end-to-end approaches built on MLLMs are often pretrained with reasoning tasks in 2D image space rather than the native 3D space in which autonomous vehicles plan.
To this end, we propose \ourmethod{}, a \underline{s}calable \underline{s}elf-\underline{s}upervised motion planning algorithm with \underline{s}patio-temporal visual representation,  based on the popular PaLI~\cite{chenpali} multimodal large language model. \ourmethod{}\ uses a novel sparse volume strategy to seamlessly transform the strong visual representation of MLLMs from perspective view to 3D space without the need to finetune the vision encoder. This representation aggregates multi-view and multi-frame visual inputs and enables better prediction of planning trajectories in 3D space.
To validate our method, we run experiments on both nuScenes and Waymo Open Motion Dataset (with in-house camera data).
Results show that \ourmethod{} performs favorably against existing supervised multi-task approaches while requiring no human annotations. It also demonstrates great scalability when pretrained on large volumes of unannotated driving logs.
\end{abstract}

\section{Introduction}
\label{sec:intro}
The exploration of end-to-end autonomous driving dates back to the 1980s~\cite{pomerleau1988alvinn}, where the motion planning model directly predicts control signals or future waypoints based on raw sensor inputs. Due to limited robustness, these early attempts struggle in complex urban situations. Recent neural network capability advances have led to renewed interest in this field~\cite{hu2023planning,jiang2023vad,weng2024drive}, especially given the strong generalization ability of Multimodal Large Language Models (MLLMs)~\cite{chen2023pali,team2023gemini,gpt4v}. 


\begin{figure}[t!]
\centering
    \centering
    \begin{subfigure}{0.85\linewidth}
        \centering
        \includegraphics[width=\linewidth]{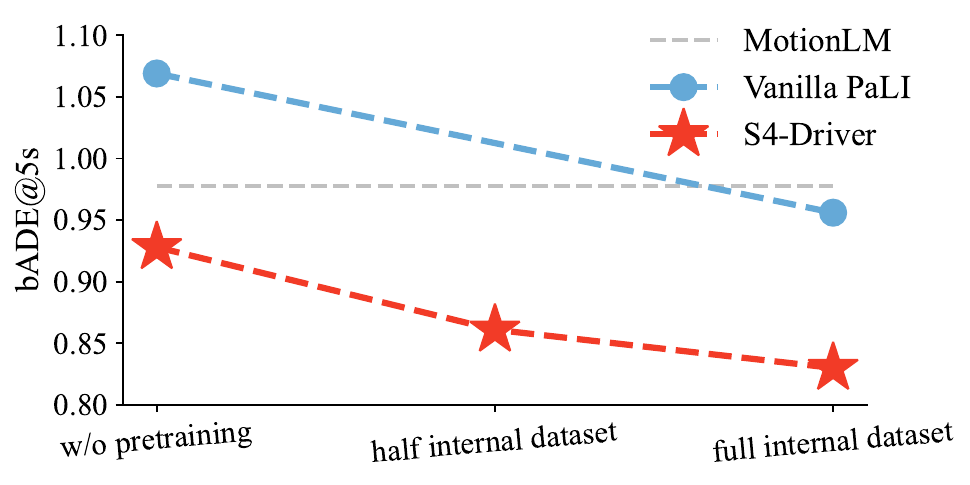}
        \vspace{-18pt}
        \caption{Pretraining data scaling-up of \ourmethod{}.}
        \label{fig:scale-up}
    \end{subfigure}
    \begin{subfigure}{0.48\linewidth}
        \centering
        \includegraphics[width=\linewidth]{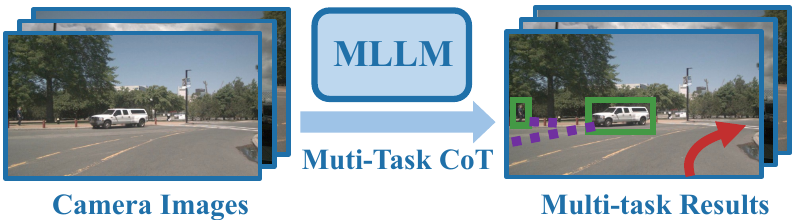}
        \caption{Multi-task learning framework.}
        \label{fig:mtl}
    \end{subfigure}
    \hfill
    \begin{subfigure}{0.48\linewidth}
        \centering
        \includegraphics[width=\linewidth]{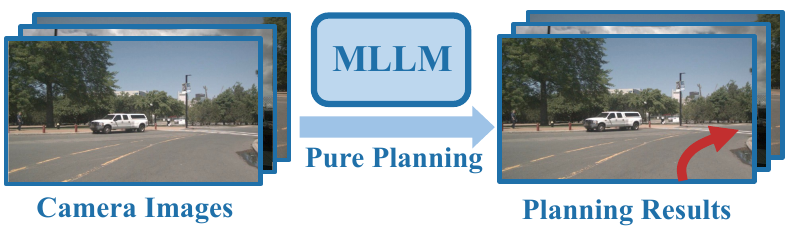}
        \caption{Self-supervised framework.}
        \label{fig:selfsup}
    \end{subfigure}
    \vspace{-10pt}
\caption{{\ourmethod} scaling up with abundant unlabeled driving logs (\ref{fig:scale-up}). Different from past efforts that improve MLLM planning via multi-task learning and CoT reasoning (\ref{fig:mtl}), \ourmethod{} focuses on self-supervised motion planning (\ref{fig:selfsup}) without human annotations. While a vanilla PaLI approach with self-supervised training achieves good performance, the full {\ourmethod} with spatio-temporal visual representation achieves the best performance and scalability (\ref{fig:scale-up}).
}
\label{fig:framework}
\vspace{-15pt}
\end{figure}

However, vanilla application of MLLMs to end-to-end motion planning can hardly exploit their strong visual understanding and reasoning capabilities given the significant difference between motion planning and MLLM pretraining tasks, resulting in inferior planning performance~\cite{tian2024drivevlm,xu2024drivegpt4}. To narrow this gap, as shown in Fig.~\ref{fig:mtl}, previous methods resort to either \textit{multi-task learning}, which incorporates diverse perception and prediction tasks into training and inference, or \textit{supervised perception pretraining}, which utilizes the pretrained autonomous driving perception models as visual tokenizers. However, fine-grained human annotations serve as the bottleneck for both strategies. In contrast, pure self-supervised approaches, while capable of learning directly from sensor inputs and utilizing the abundant unlabeled driving logs, often under-perform the state of the art.

To challenge this status-quo, we aim to boost the performance of self-supervised motion planning orthogonal to prior work (Fig.~\ref{fig:selfsup}). It only requires raw sensor data and ego-vehicle trajectories for self-supervised model training, 
without the need of expensive  human labels.
First of all, we identify the following two main obstacles in this direction:


\begin{itemize}
    \item \textbf{Suboptimal representation format.} MLLMs~\cite{chen2023pali,chenpali} are typically designed for tasks in the 2D image plane. This image space representation limits their 3D reasoning ability from the combination of multi-view images. 
    \item \textbf{Limited data scale.} Although it is one of the most widely adopted benchmarks for end-to-end planning, nuScenes~\cite{caesar2020nuscenes} only contains fewer than 1k sequences and lacks driving behavior diversities. This limited scales leads to severe over-fitting issues in the finetuning of MLLMs with billion-parameter size.
\end{itemize}

In this paper, we handle the above challenges to better explore the potential of MLLMs. Firstly, we propose \textbf{\ourmethod{}}, a simple yet effective \textbf{\underline{S}}calable \textbf{\underline{S}}elf-\textbf{\underline{S}}upervised motion planning method with \textbf{\underline{S}}patio-temporal visual representation.
Built upon a general multimodal large language model~\cite{chenpali, chen2023pali} , we directly predict ego-vehicle waypoints from camera images, eliminating the need for intermediate perception and prediction tasks,
and thus facilitating the scale-up of model pretraining leveraging massive unannotated driving logs (Fig.~\ref{fig:scale-up}). To address the obstacle of suboptimal visual format, we propose a novel sparse volume representation that enables aggregating visual information from multi-view and multi-frame images. Moreover, this spatio-temporal representation is designed to work with frozen features from the perspective-view vision encoder. It boosts 3D spatio-temporal reasoning capabilities of the model on motion planning, and seamlessly preserves world knowledge in the pretrained visual embeddings of MLLMs. 
Secondly, to rigorously evaluate our method with sufficient training data, we also utilize large-scale \ourdataset{} benchmark~\cite{ettinger2021large} with in-house camera sensor data besides the popular nuScenes dataset~\cite{caesar2020nuscenes}. \ourdataset{} is approximately $100\times$ larger than nuScenes, so it serves as a more comprehensive benchmark to evaluate model performance in a scalable manner.

In summary, our contributions include:

\begin{itemize}
   \item We advocate for a simple self-supervised end-to-end motion planning with MLLMs, solely relying  on ego-vehicle trajectory supervision. This approach simplifies the system architecture and facilitates data scaling for training.
    \item We propose a novel sparse volume representation to effectively aggregate multi-view and multi-frame visual information, enhancing the 3D spatio-temporal reasoning capabilities of the MLLM for motion planning.
    \item Our algorithm achieves state-of-the-art performance on both nuScenes and {\ourdataset} benchmarks. It also  demonstrate great scalability with pretraining.
    
\end{itemize}

\section{Related Works}
\textbf{Multimodal Large Language Models (MLLMs).} MLLMs with both language and images modalities have garnered significant attention in recent years~\cite{radford2021learning,alayrac2022flamingo,li2023blip,achiam2023gpt,team2023gemini}. Prior research has focused on integrating powerful Large Language Models (LLMs)~\cite{zheng2023judging,chung2024scaling,zheng2023judging} with advanced image encoders~\cite{fang2023eva,radford2021learning,zhai2023sigmoid} such as LLaVA~\cite{liu2024visual}, PaLI~\cite{chenpali,chen2023pali}, PaliGemma~\cite{beyer2024paligemma}, and InstructBLIP~\cite{instructblip}. Through instruction tuning or multi-modal finetuning,
these models demonstrate impressive performance in multimodal understanding and reasoning. Current developments are leveraging increasingly larger multimodal datasets to further enhance their capabilities in complex perception and generalization tasks~\cite{bai2023qwen,gpt4v,grok}. However, despite their strengths, these models have shown limitations in 3D spatial reasoning~\cite{hong20233d,cheng2024spatialrgpt,mckinzie2024mm1}, posing challenges for their application in autonomous driving.


\textbf{End-to-end Autonomous Driving.} While the key final step of autonomous driving is planning, traditional systems are often assembled with multiple modules for perception~\cite{liu2023bevfusion,xie2023sparsefusion,liaomaptr} and prediction~\cite{varadarajan2022multipath++,nayakanti2023wayformer,zheng2023occworld} tasks. To mitigate the information loss and error accumulation across modules, end-to-end driving systems~\cite{casas2021mp3,hu2022st,hu2023planning,jiang2023vad,zheng2024genad,li2024ego} utilize a unified model to predict the ego-vehicle future waypoints or control signals directly from raw sensor inputs. While prioritizing planning, these systems typically still incorporate perception and prediction, requiring explicit supervision for each.  Although some prior works~\cite{seff2023motionlm,codevilla2018end,wu2022trajectory} have explored pure motion planning without any intermediate tasks, they struggle in complex urban scenarios due to modeling capability limitations.

\textbf{MLLMs for Driving.} The superior reasoning and generalization abilities of large-scale models are desirable properties for autonomous driving. Some efforts transform the driving scenario into large language model textual prompts~\cite{sha2023languagempc,mao2023gpt,chen2024driving} or directly process camera images with vision-language models~\cite{xu2024drivegpt4,shao2024lmdrive,ding2024holistic}. However, their potential is limited to the small scale of existing benchmarks~\cite{caesar2020nuscenes}, which only allow partial finetuning~\cite{hu2021lora}. Meanwhile, closed-loop simulators~\cite{dosovitskiy2017carla} struggle to provide realistic sensor data for end-to-end tasks. Consequently, multi-task joint finetuning~\cite{simadrivelm,tian2024drivevlm,xu2024drivegpt4} or chain-of-though (CoT)~\cite{wei2022chain} inference~\cite{ma2023dolphins,mao2023language,mao2023gpt} are widely adopted to simplify the reasoning despite the extra annotation requirements brought by perception and prediction modules. Alternatively, some work~\cite{tiantokenize,ding2024holistic} integrates pretrained perception models~\cite{li2022bevformer} to extract BEV features and sends to language models as vision tokens. Recently, EMMA~\cite{hwang2024emmaendtoendmultimodalmodel} leverages powerful Gemini~\cite{team2023gemini} for self-supervised motion planning. Additionally, they develop a mixture of training tasks, including motion planning, 3D object detection, and road graph estimation as well as additional reasoning process for trajectory generation. In contrast, our work focuses on the enhancement of self-supervised motion planning without extra human labels.

\section{Enhancing MLLM for Motion Planning}

\begin{figure*}[t]
    \centering
    \includegraphics[width=0.95\linewidth]{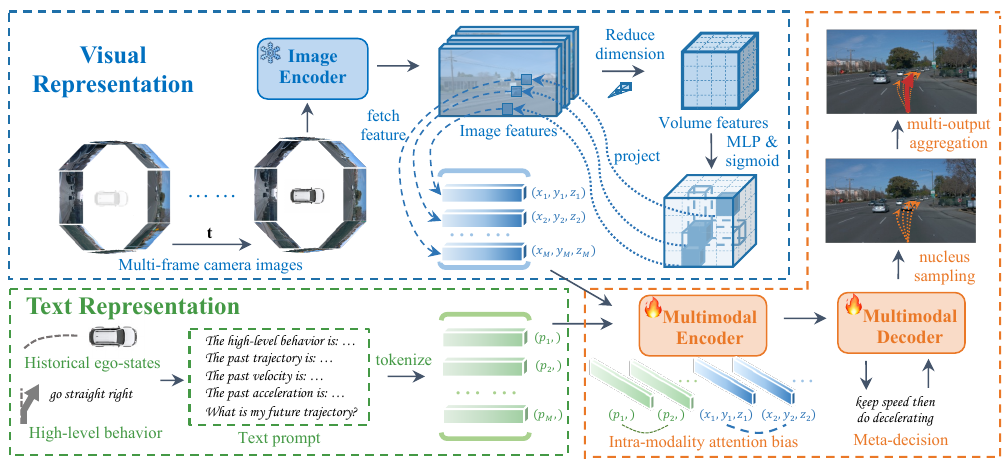}
    \vspace{-8pt}
    \caption{Overview of our proposed \ourmethod{} algorithm. We enhance the PaLI model for motion planning by incorporating meta-decision, spatio-temporal visual representation, and multi-decoding aggregation.
    }
    \label{fig:overview}
    \vspace{-10pt}
\end{figure*}

\begin{figure}[t]
    \vspace{-4 pt}
    \centering
    \includegraphics[width=0.9\linewidth]{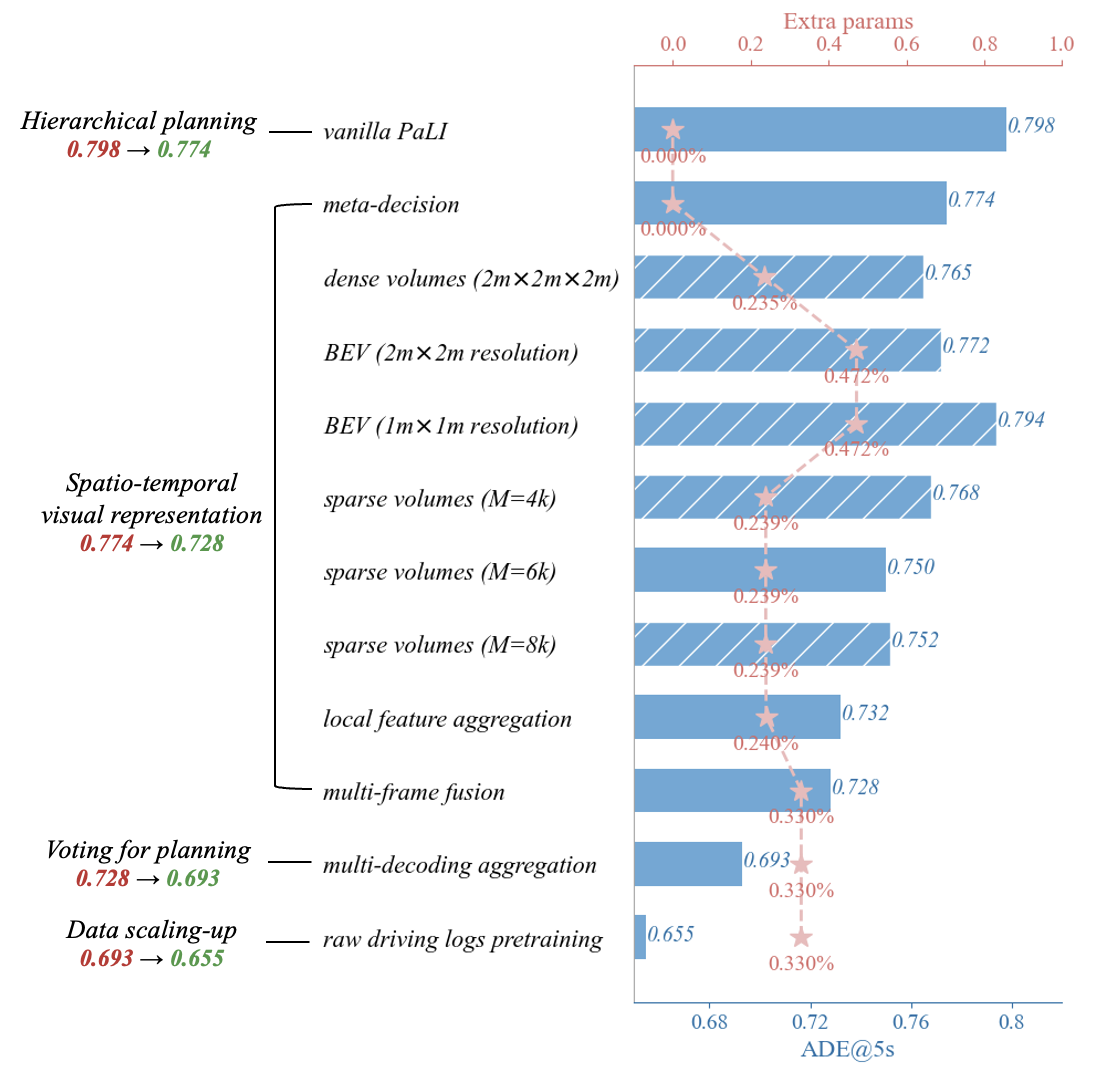}
    \vspace{-10pt}
    \caption{A roadmap for enhancing MLLM for planning. The techniques are adopted step-by-step from top to bottom, while shadow items are not adopted in the subsequent steps. We show the performance on \ourdataset{} (Sec.~\ref{sec:benchmark}) after including each module.}
    \label{fig:roadmap}
    \vspace{-18pt}
\end{figure}

In this section, we adapt a pretrained MLLM for the end-to-end motion planning task. Without loss of generality, we leverage PaLI~\cite{chen2023pali}. However, these techniques are also compatible with other more powerful MLLMs. We introduce a set of design improvements to the MLLM setup and demonstrate correspondingly improved performance on \ourdataset{} (Fig.~\ref{fig:roadmap}).


\subsection{Vanilla PaLI as Planner}
\label{sec:vanilla}
End-to-end motion planning models determine the future trajectory $\mathcal{O}_{T_f}$ of ego-vehicle based on multi-view camera images $\mathcal{C}$ and high-level behavior command $b$. The future trajectory consists of the ego-vehicle's location at each future time step in the BEV coordinate, \textit{i.e.} $\mathcal{O}_{T_f}=\left[(x_t,y_t)\right]_{t=1}^{T_f}$. The high-level behavior command is necessary to guide the driving direction, similar to a navigation system. Additionally, as verified in prior works~\cite{zhai2023ADMLP,li2024ego}, the historical states of ego-vehicle $\mathcal{H}_{T_h}$ are also critical for a smooth and feasible planning result, where we consider the historical locations, velocities, and accelerations as $\mathcal{H}_{T_h}=\left[\mathbf{l}_t,\mathbf{v}_t,\mathbf{a}_t\right]_{t=-1}^{-T_h}$, \textit{i.e.}
\begin{equation}
    \mathcal{O}_{T_f} = P(\mathcal{C}, \mathcal{H}_{T_h}, b)
\end{equation}
where $P(\cdot)$ is the planning model. We provide both historical ego states and high-level commands to the model as text prompts. Locations, velocities, and accelerations are represented directly as floating numbers with two decimals. The predicted future trajectories are then extracted from the decoded text outputs of the model. Example prompts and targets are provided in the supplementary materials. 

Without preceding perception and prediction tasks, the vanilla PaLI, fine-tuned in a self-supervised manner, achieves reasonable but not ideal performance in motion planning (Fig.~\ref{fig:roadmap}). 


\subsection{Hierarchical Planning with Meta-Decision}
\label{sec:decision}
Directly outputting a future trajectory without any reasoning is challenging for MLLMs. To address this, we apply a coarse-to-fine approach, inspired by Chain-of-Thought (CoT)~\cite{wei2022chain}. It adopts a hierarchical planning method~\cite{igl2022symphony}, going from from semantic decision to numeric planning. We prompt our model to provide a high-level estimation of the future acceleration state $D$ first, effectively decomposing the motion planning task into two steps:
\begin{equation}
     \mathcal{D} = P_{s1}(\mathcal{C}, \mathcal{H}_{T_h}, b),\quad \mathcal{O}_{T_f} = P_{s2}(\mathcal{C}, \mathcal{H}_{T_h}, b;\mathcal{D})
\end{equation}
We define $D$ to encompass four meta-decisions: \textit{keep stationary, keep speed, accelerate} and \textit{decelerate}.  Notably, unlike previous CoT applications in VLM-based planning that require human annotation for training, such as the scene analysis in DriveVLM~\cite{tian2024drivevlm}, we incorporate these meta-decisions as a ``free lunch" to simplify the motion planning process without requiring any additional annotations.
The ground-truth decisions are generated by heuristic rules based on the future ego velocity and acceleration (details in the supplementary materials). Fig.~\ref{fig:roadmap} shows this simple design brings non-trivial improvements in planning performance.

\subsection{Scene Representation in 3D Space}
\label{sec:3dscene}
High-quality motion planning requires a robust understanding of the surrounding 3D scene, including both static and dynamic elements. While traditionally achieved through separate perception and prediction modules, our self-supervised end-to-end framework relies on the MLLM to implicitly learn this understanding without explicit supervision. However, despite strong 2D reasoning capabilities, MLLMs often struggle with 3D spatial reasoning due to limitations in their perspective view representation and the lack of depth-related tasks in pretraining~\cite{chen2024spatialvlm, ma2024llms}.


\subsubsection{3D Visual Representation with Dense Volumes}
To overcome the above limitations, we draw inspiration from previous successful perception tasks~\cite{li2022bevformer,liu2023bevfusion} and adopt a 3D volume representation. The multi-view feature maps $\mathcal{F}_{2D}=\{\mathbf{f}^v_{2D}\}_{v=1}^V (\mathbf{f}^{v}_{2D}\in\mathbb{R}^{H\times W\times C})$ are extracted by visual encoder of MLLM, where $V$ is the number of views. We construct a 3D feature volume $\mathbf{f}_{3D}^{vol}\in\mathbb{R}^{X\times Y\times Z\times C}$ centered at the ego-vehicle based on the multi-view image features. 
To avoid introducing complex modules that might disrupt the pretrained MLLM and misalign visual features with the subsequent multimodal encoder-decoder, we employ a lightweight projection method, similar to SimpleBEV~\cite{harley2023simple}.  Specifically, for each voxel in the 3D volume, we  project its  $(x,y,z)$ coordinates  to each perspective view $v$, obtaining corresponding 2D coordiantes $(u_v,v_v)$. We then bilinearly sample the local features from each view at these projected locations. Finally, the voxel's feature representation is computed as the average of the local semantic features from all views where the voxel projects within the image bounds. This process effectively incorporates 3D spatial information while maintaining compatibility with the pretrained MLLM.
\begin{gather}
    \mathbf{f}_{3D,(x,y,z)}^{vol}=\mathbf{f}_{(x,y,z)}^{sem} + \text{PosEmb}(x,y,z)     \label{eq:projection}\\    
    \mathbf{f}_{(x,y,z)}^{sem}= \text{Avg}\left(\left\{\text{Bilinear}\left(\mathbf{f}^v_{2D};(u_v,v_v)\right)\right\}_v\right)
    \label{eq:bilinear}
\end{gather}

This simple and efficient projection strategy ensures that the 3D volume features $\mathbf{f}_{3D}^{vol}$ share a similar distribution with the original multi-view features $\mathcal{F}_{2D}$. This similarity facilitates seamless integration with the subsequent multimodal encoder-decoder of the MLLM. 

As shown in Fig.~\ref{fig:roadmap}, this 3D volume representation leads to some improvement in motion planning performance. Alternatively, we find that reducing the $Z$-axis using a fully connected layer to obtain a BEV representation yields slightly worse performance since this reduction operation may introduce ambiguity to the scene representation.

\subsubsection{Sparse Volume Representation}
\label{sec:sparse}
While the 3D volume representation effectively captures spatial information, 
inherently most of the surrounding 3D space is empty.
Moreover, for motion planning, detailed information about objects distant from the road, such as buildings and trees, is less critical. Based on this observation, we propose a sparse volume representation to reduce the number of voxels, enabling higher resolution given memory constraints and improving efficiency.

To determine useful volumes in each scene based on their positions and semantics, we define a gate value $g^{vol}_{(x,y,z)}\in(0,1)$ for each volume at coordinates $(x,y,z)$. To obtain this gate, starting from multi-view image features $\left\{\mathbf{f}_{2D}^v\right\}_{v=1}^V$, we reduce its dimension through a fully connect (FC) layer as 
\begin{equation}
    \mathbf{f}_{2D}^{gate,v}=\text{FC}(\mathbf{f}_{2D}^v),\quad \mathbf{f}_{2D}^{gate,v}\in\mathbb{R}^{H\times W\times C'},C'\ll C.
    \label{eq:reduce_dim}
\end{equation}
Then we construct a reduced-dimension volume feature $\mathbf{f}^{gate}_{3D}\in\mathbb{R}^{X\times Y\times Z\times C'}$ from  $\{\mathbf{f}_{2D}^{gate,v}\}_{v=1}^V$ same as Eq.~\ref{eq:projection}. A smaller channel number allows for larger volume resolution and is sufficient to indicate whether a volume is related to motion planning . Afterward, the gate is derived from $\mathbf{f}^{gate}_{3D}$ through a small MLP module as follows.
\begin{equation}
    g^{vol}_{(x,y,z)} = \text{sigmoid}\left(\text{MLP}(\mathbf{f}^{gate}_{3D,(x,y,z)})\right)
\end{equation}
Thus, we can easily select $M$ volumes ($M\ll X\cdot Y\cdot Z$) with largest gate values with coordinates $\{(x_i,y_i,z_i)\}_{i=1}^M$.


As we do not have access to ground-truth occupancy state, we propose learning the gate values implicitly. We assume that regions with small gate values should be empty or irrelevant to planning. For these vacant space, we assign a learnable feature $\mathbf{f}_{vac}\in\mathbb{R}^C$. We expect the model can learn the gate value via  the trade-off between the semantic features and this vacant feature at each 3D position, so we fetch features for selected sparse volumes as
\begin{gather}
\mathbf{f}_{3D,(x,y,z)}^{sparse}=\mathbf{\hat{f}}_{3D,(x,y,z)}^{sparse} + \text{PosEmb}(x,y,z) \label{eq:sparse_project}\\
\mathbf{\hat{f}}_{3D,(x,y,z)}^{sparse} = g_{(x,y,z)}^{vol}\cdot \mathbf{f}_{(x,y,z)}^{sem} + (1-g_{(x,y,z)}^{vol})\cdot \mathbf{f}_{vac} \label{eq:vac}
\end{gather}
where $\mathbf{f}_{(x,y,z)}^{sem}$ is same as Eq.~\ref{eq:bilinear} and $(x,y,z)\in\{(x_i,y_i,z_i)\}_{i=1}^M$ are selected volumes with large gate values. When the sparse volume features $\mathbf{f}_{3D,(x,y,z)}^{sparse}$ are fed into the subsequent multimodal encoder, they notably boost the planning performance (Fig.~\ref{fig:roadmap}) by simultaneously providing explicit 3D spatial cues of surrounding scenes and concentrate on the important regions adaptively.

\subsubsection{Local Feature Aggregation in 3D Space} 
\label{sec:bias}
Due to the lack of depth information, the lifting process in Eq.~\ref{eq:projection} or \ref{eq:sparse_project} results in duplicate volume features along each camera ray. This spatial ambiguity can be mitigated by 3D local operations like convolutions~\cite{harley2023simple} or deformable attention~\cite{li2022bevformer} as demonstrated in previous work. However, in our MLLM framework, it is less economical to insert extra local operations. Instead, we inject some relative position bias into the existing multimodal encoder by tailoring the self-attention as follows.
\begin{equation}
    \text{Attn}(Q,K,V)=\text{Softmax}\left(\frac{QK^T}{\sqrt{d}}-b(D)\right)V
    \label{eq:bias}
\end{equation}
Given $M$ sparse volumes with coordinates $\{(x_i,y_i,z_i)\}_{i=1}^M$, the distance matrix $D\in\mathbb{R}^{M\times M\times 3}$ is the distance along $xyz$-axes between each pair of sparse volumes, \textit{i.e.} $D_{ij}=(x_j-x_i,y_j-y_i,z_j-z_i)$. The bias is computed with the function $b(\Delta x,\Delta y, \Delta z)=b_x(\Delta x)+b_y(\Delta y)+b_z(\Delta z)$, where $\Delta x,\Delta y,\Delta z$ are divided into several bins and mapped to a learnable bias value for each bin through $b_x(\cdot),b_y(\cdot),b_z(\cdot)$.  We also apply a separate bias between the text tokens with 1D position $\{p_i\}$.

This relative position bias elegantly inserts local inductive bias to pretrained global self-attention modules with minimal additional costs. It leads to a notable performance gain in Fig.~\ref{fig:roadmap} by facilitating the aggregation of local information in 3D space, enhancing scene understanding and spatial reasoning.

\subsection{Multi-frame Temporal Fusion}
\label{sec:temporal}
Multi-frame input combination helps to compensate for the lack of depth cues in camera images. We extend our sparse volume representation to aggregate mult-frame temporal information by incorporating $T$ historical frames with an interval of $0.5s$. Given images from the total $T+1$ frames, Eq.~\ref{eq:reduce_dim} is applied to each frame separately to obtain the multi-view image features $\mathbf{f}_{2D,t}^{gate,v}, t=0,-1,\dots,-T, v=1,\dots, V$. After ego-motion compensation, we construct a gate feature volume $\mathbf{f}_{3D,t}^{gate}$ based on the current ego-vehicle coordinates and per-frame image features. Multi-frame volume features are concatenated along the channel dimension to generate gate value.
\begin{equation}
    g^{vol}_{(x,y,z)} = \text{sigmoid}\left(\text{MLP}\left(\text{Concat}\left([\mathbf{f}^{gate}_{3D,t,(x,y,z)}]_{t=0}^{-T}\right)\right)\right)
\end{equation}
Similar with Sec.~\ref{sec:sparse}, we select $M$ volumes based on the gate value. Following Eq.~\ref{eq:sparse_project}, the volume features are fetched from each frame image feature map separately as $\mathbf{f}^{sparse}_{3D,t,(x,y,z)},t=0,\dots,-T$, which are fused with an FC-layer as the sparse volume feature with temporal awareness.
\begin{equation}
    \mathbf{f}^{sparse}_{3D,(x,y,z)}=\text{FC}\left(\text{Concat}\left(\left[\mathbf{f}^{sparse}_{3D,t,(x,y,z)}\right]_{t=0}^{-T}\right)\right)
\end{equation}
In Fig.~\ref{fig:roadmap}, temporal fusion helps to improve motion planning performance by facilitating the environment understanding.

\subsection{Voting for Planning via Multi-Decoding}
\label{sec:multi}
MLLMs are prone to assigning high confidence to the simple future behaviors in motion planning, \textit{e.g.}, keeping stationary. To relieve this bias, we aggregate multiple outputs and use voting to obtain the final planning output. We resort to nucleus sampling~\cite{holtzman2019curious} to produce multiple future trajectories for the ego-vehicle, denoted as $\{\mathcal{O}_{T_f}^k\}_{k=1}^K$. They are simply averaged to produce a unique planning result as follows: $\mathcal{O}_{T_t}=\left[(x_t,y_t)\right]_{t=1}^{T_f}$.
\begin{equation}
   \quad (x_t,y_t)=\left(\frac{1}{K}\sum_{k=1}^{K}x_t^k, \frac{1}{K}\sum_{k=1}^{K}y_t^k\right)
\end{equation}
This unweighted average mitigates the MLLM's bias towards simple behaviors. Fig.~\ref{fig:roadmap} demonstrates the notable performance gain with this simple multi-decoding aggregation approach.

\subsection{Scaling to Large-scale Raw Driving Logs}
The self-supervised training allows our proposed \ourmethod{} to scale to large-scale driving logs amounts without requiring human annotations. To exploit the potential of MLLM-based planners, we pretrain the model on our internal dataset. Results in Fig.~\ref{fig:roadmap} demonstrate that \ourmethod{} achieves notable performance gains in challenging tail behaviors due to large-scale pretraining.
\begin{table}[t!]
    \centering
    \resizebox{\linewidth}{!}{
    \begin{tabular}{c|cc|cc|cc}
        \toprule
         \multirow{2}{*}{\textbf{Benchmark}} & \multicolumn{2}{c|}{\textbf{Size}} & \multicolumn{2}{c|}{\textbf{Data Format}} & \multicolumn{2}{c}{\textbf{Metrics}}\\
         & Sequence & Length & Frequency & Planning horizon & Sample-wise & Behavior-wise  \\
         \midrule 
         BDD-X~\cite{kim2018textual} & 7k & 77 h & 10 Hz & control signal & \Checkmark  & \XSolidBrush\\
         nuScenes~\cite{caesar2020nuscenes} & 1k & 5.5 h & 2 Hz & 3s & \Checkmark& \XSolidBrush  \\
         \midrule
         \ourdataset{} & 103k & 574 h & 10 Hz & 5s & \Checkmark & \Checkmark \\
        \bottomrule
    \end{tabular}
    }
    \vspace{-10pt}
    \caption{Comparisons between \ourdataset{} and existing counterparts. \ourdataset{} has a significantly larger size and more comprehensive behavior-wise metrics.}
    \label{tab:dataset}
\end{table}

\begin{figure}[t!]
    \centering
    \includegraphics[width=0.9\linewidth]{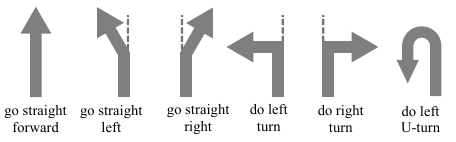}
    \vspace{-8 pt}
    \caption{High-level behavior commands for motion planning.}
    \label{fig:commands}
\end{figure}

\section{Waymo Open Motion Dataset for Planning}
\label{sec:benchmark}
For the large-scale training and evaluation of planning algorithms with large models, we design a \ourdataset{} benchmark based on WOMD dataset~\cite{ettinger2021large}, and compare with the existing datasets in Tab.~\ref{tab:dataset}.

\begin{figure}[t!]
    \centering
    \begin{subfigure}{0.18\linewidth}
        \centering
        \includegraphics[width=\linewidth]{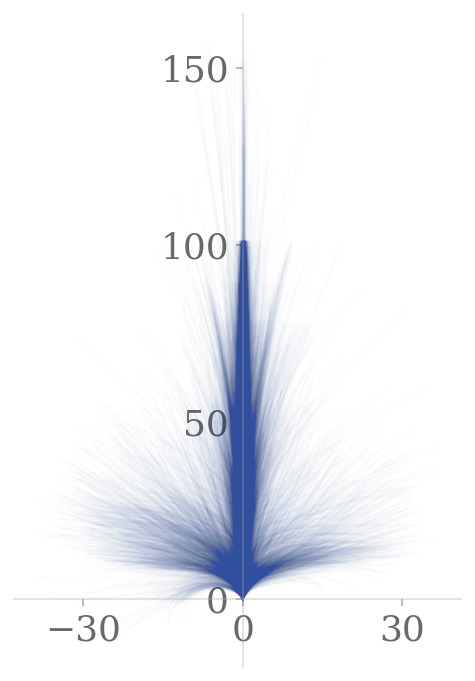}
        \vspace{-15pt}
        \caption{Trajectory}
    \end{subfigure}
    \hfill
    \begin{subfigure}{0.26\linewidth}
        \centering
        \vspace{9pt}
        \includegraphics[width=\linewidth]{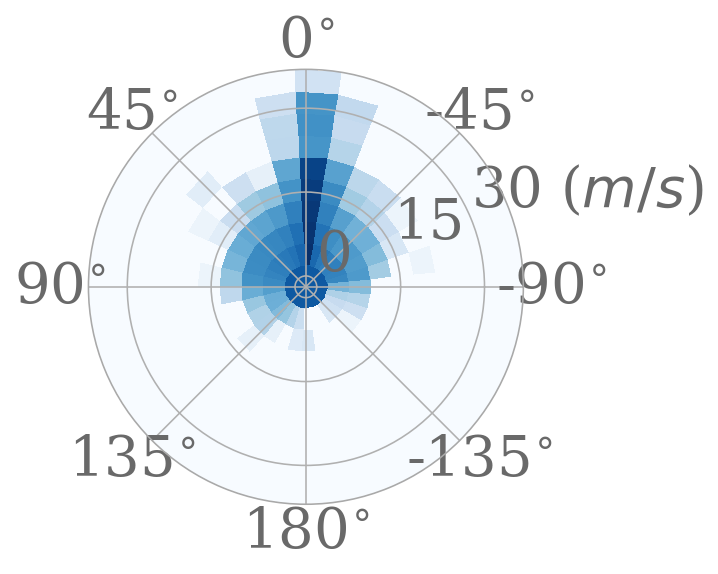}
        \caption{Velocity}
        \label{fig:velocity}
    \end{subfigure}
    \hfill
    \begin{subfigure}{0.28\linewidth}
        \centering
        \vspace{9pt}
        \includegraphics[width=\linewidth]{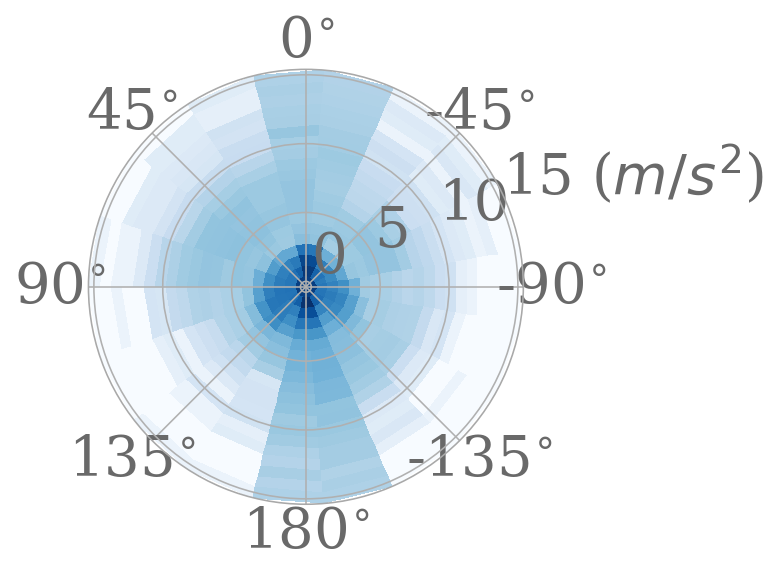}
        \caption{Acceleration}
        \label{fig:acceleration}
    \end{subfigure}
    \hfill
    \begin{subfigure}{0.25\linewidth}
        \centering
        \includegraphics[width=\linewidth]{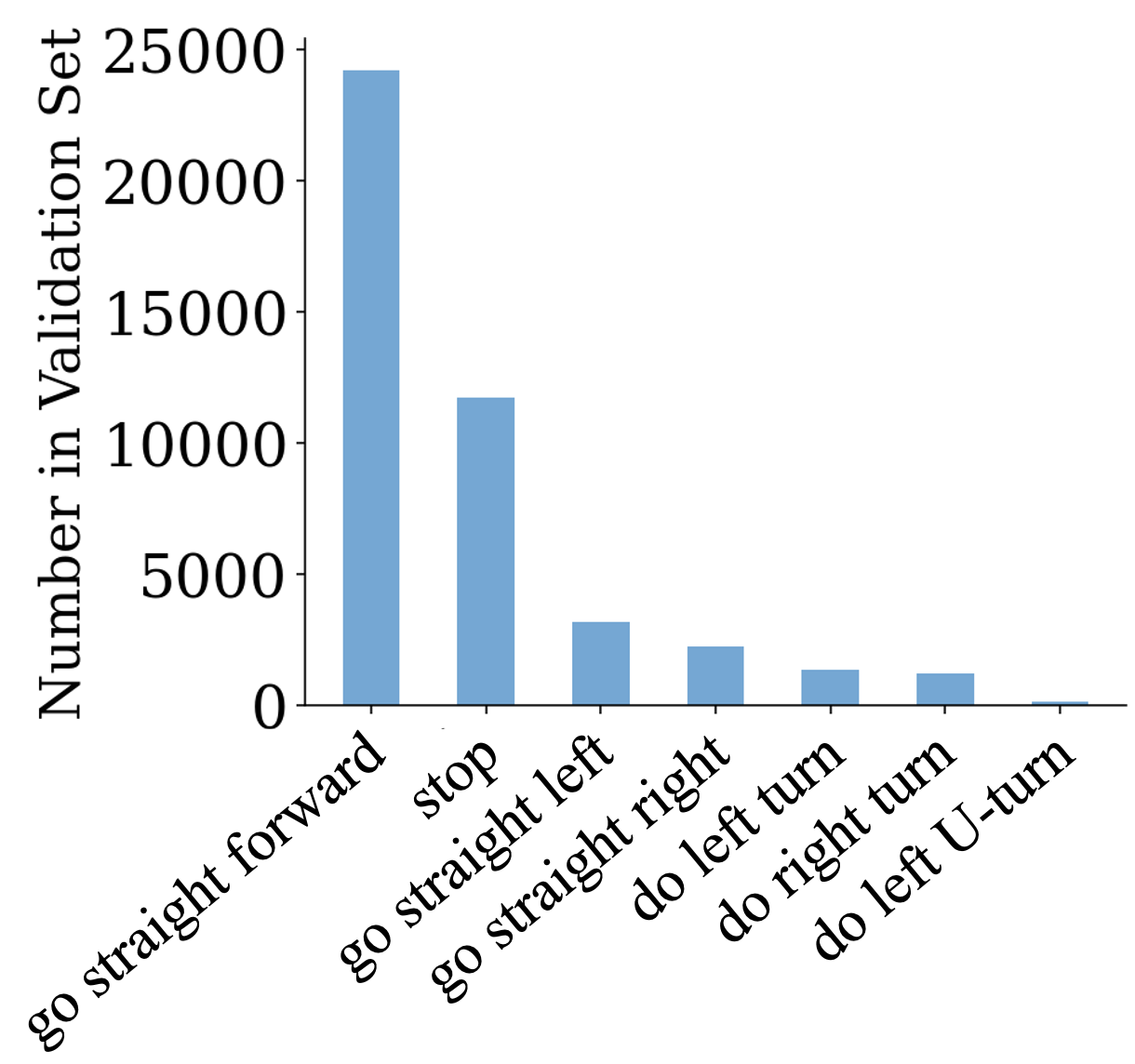}
        \caption{Behavior}
    \end{subfigure}
    \caption{Statistical distributions of \ourdataset{}. We show the distributions of ego-vehicle trajectories, velocities, accelerations and each behavior scenario sizes in the validation set.
    }
    \label{fig:distribution}
    \vspace{-10pt}
\end{figure}

\noindent\textbf{Overview.} It contains $103k$ real-world driving scenarios, covering diverse urban and suburban scenes. They are further divided into $9s$ samples with $1s$ history and $8s$ future. For end-to-end planning evaluation,  besides the ego-vehicle trajectories in each sample as groundtruth, we also have the following key items in the dataset:

\noindent\textbf{(In-house) Camera Data.} 
Most end-to-end planning methods~\cite{jiang2023vad,simadrivelm,xu2024drivegpt4,tian2024drivevlm} relies on the camera images as model inputs. In our dataset, each frame contains images captured by eight multi-view cameras.

\noindent\textbf{High-level Behavior Commands.} Like navigation systems for humans (\textit{e.g.} Google Map), the end-to-end planning system also requires navigation signals to indicate the direction to go. We consider six high-level behavior commands (Fig.~\ref{fig:commands}), namely \textit{go straight forward, go straight left, go straight right, do left turn, do right turn, do left U-turn}. They can cover diverse driving situations in real world, \textit{e.g.} \textit{``go straight right"} describes the case to leave the high-way. Instead of simply considering the last step position~\cite{casas2021mp3,hu2023planning}, we decide the behavior commands based on long-term future trajectories, which can handle the low-speed or stopping situations. Details of this heuristic rule are included in the supplementary materials.

\noindent\textbf{Evaluation Metrics.} The imbalance of data distribution is inevitable in real driving scenarios. For example, in the \ourdataset{} benchmark, moving straight and stopping account for more than $70\%$ of all the samples, as reflected by Fig.~\ref{fig:distribution}. In this case, we argue that current widely used sample-wise average displacement error and collision rates cannot comprehensively reflect the performance of motion planning algorithms since challenging but less frequent behaviors, like turning, are overwhelmed by simple straightforward moving scenarios. To this end, we propose novel behavior-wise metrics similar to the mAP metric in prediction~\cite{ettinger2021large}. For example, we denote the behavior-wise average displacement error as \textit{bADE} defined as follows.
\begin{equation}
    bADE=\frac{1}{|\mathcal{B}|}\sum_{b\in\mathcal{B}}ADE_b
\end{equation}
where $ADE_b$ is the ADE metric for a specific behavior. Concretely, $|\mathcal{B}|=7$ behaviors are considered -- including the six high-level commands (Fig.~\ref{fig:commands}) and an additional \textit{stop} behavior\footnote{We do not include ``stop" in the high-level behavior commands since navigation system cannot provide it and this may leak future information like traffic lights. Details are explained in supplementary materials.}. These behavior-wise metrics make up for the shortcomings of previous sample-wise metrics by paying more attention to those scarce but safety-critical behaviors.

\section{Experiments}
\subsection{Implementation Details}
\noindent\textbf{Model and finetuning.} Unless otherwise specified, we construct our model from a pretrained PaLI3-5B model~\cite{chen2023pali}, which includes a ViT-G (2B) vision encoder~\cite{dosovitskiy2020image} and multimodal encoder-decoder (3B)~\cite{tay2022ul2}. Our default setting freezes the ViT encoder and finetune only the inserted modules and the multimodal encoder-decoder. Finetuning the model on the \ourdataset{} dataset requires approximately 2.5 days on 128 Google Cloud TPU v4 using a batch size of 256 and a learning rate of 3e-3.

\noindent\textbf{Datasets.} We evaluate \ourmethod{} on both \ourdataset{} (Sec.~\ref{sec:benchmark}) and nuScenes~\cite{caesar2020nuscenes} benchmarks. Given the limited size of nuScenes, which is insufficient for training a multimodal large language model, we also attempt to finetune the model on nuScenes using weights pretrained on \ourdataset{}.

\begin{table*}[ht]
    \centering
    \begin{tabular}{c|cc|cc|cccc}
    \toprule
      \multirow{2}{*}{\textbf{Methods}} & \multicolumn{2}{c|}{\textbf{Human Annotations}} & \multicolumn{2}{c|}{\textbf{Extra Data}} & \multicolumn{4}{c}{$L_2 (m)$} \\
      & perception & prediction &data & labels& 1.0s & 2.0s & 3.0s & $Avg_{1,2,3s}$ \\
     \midrule
     \multicolumn{9}{c}{\textit{Multi-task end-to-end approaches}}\\
     \midrule
     UniAD~\cite{hu2023planning}  & \Checkmark & \Checkmark &\XSolidBrush &\XSolidBrush & 0.42 & 0.64 & 0.91 & 0.66\\
     VAD~\cite{jiang2023vad}  & \Checkmark & \Checkmark &\XSolidBrush &\XSolidBrush & 0.17 & 0.34 & 0.60 & 0.37\\
     PARA-Drive~\cite{weng2024drive}  & \Checkmark & \Checkmark &\XSolidBrush &\XSolidBrush & 0.25 & 0.46 & 0.74 & 0.48\\
     \midrule
     \multicolumn{9}{c}{\textit{Multimodal large language model approaches}}\\
     \midrule
     GPT-Driver~\cite{mao2023gpt} &  \Checkmark & \Checkmark & \XSolidBrush &\XSolidBrush  & 0.20 & 0.40 & 0.70 & 0.44\\
     DriveVLM~\cite{tian2024drivevlm}  &  \Checkmark & \Checkmark & \Checkmark & \Checkmark & 0.18 & 0.34 & 0.68 & 0.40\\
     OmniDrive~\cite{wang2024omnidrive}  &  \Checkmark & \Checkmark & \Checkmark & \Checkmark& 0.14 & 0.29 & 0.55 & 0.33\\
     \midrule
     \multicolumn{9}{c}{\textit{Self-supervised approaches}}\\
     \midrule
     DriveVLM (w/o CoT)~\cite{tian2024drivevlm} &  \XSolidBrush & \XSolidBrush & \Checkmark & \XSolidBrush& 0.19 &0.41& 0.89 & 0.49\\
     \ourmethod{} (ours)&  \XSolidBrush &  \XSolidBrush & \XSolidBrush &  \XSolidBrush & 0.16 & 0.34 & 0.63 & 0.38\\
     \ourmethod{}$^*$ (ours)&  \XSolidBrush &  \XSolidBrush & \Checkmark &  \XSolidBrush & \textbf{0.13} & \textbf{0.28} & \textbf{0.51} & \textbf{0.31}\\
     \bottomrule
    \end{tabular}
    \vspace{-8pt}
    \caption{NuScenes benchmark results. Most previous works require human annotations as additional supervision. \ourmethod{}$^*$ is pretrained on \ourdataset{}, while many prior MLLM-based methods also utilize extra data and labels due to the limited scale of nuScenes.}
    \label{tab:nuscenes}
    \vspace{-5pt}
\end{table*}

\begin{table*}[ht]
    \centering
    \begin{tabular}{c|ccc|ccc}
        \toprule
        \multirow{2}{*}{\textbf{Methods}} &  \multicolumn{3}{c|}{\textbf{Sample-wise Metrics}} & \multicolumn{3}{c}{\textbf{Behavior-wise Metrics}} \\
        & ADE@1s & ADE@3s & ADE@5s & bADE@1s & bADE@3s & bADE@5s \\
        \midrule
        Vanilla PaLI & 0.034 & 0.277 & 0.798 & 0.051 & 0.392 & 1.069 \\
        \ourmethod{} (ours)&0.031 & 0.247 & 0.693 & 0.049& 0.350  & 0.928 \\
        \ourmethod{}* (ours) & \textbf{0.029} & \textbf{0.233} & \textbf{0.655} & \textbf{0.044} & \textbf{0.313} & \textbf{0.830}\\
        \midrule
        \multicolumn{7}{c}{\textit{\textcolor{gray}{
        Models that utilize high-quality objects, tracks, and roadgraph as inputs rather than raw camera images.
        }}}\\
        \midrule
        \textcolor{gray}{MotionLM~\cite{seff2023motionlm}} & \textcolor{gray}{0.045} & \textcolor{gray}{0.266} & \textcolor{gray}{0.697} & \textcolor{gray}{0.066} & \textcolor{gray}{0.388} & \textcolor{gray}{0.978} \\
        \bottomrule
    \end{tabular}
    \vspace{-8pt}
    \caption{\ourdataset{} benchmark results. ``*" denotes methods pretrained on the internal dataset.}
    \label{tab:womd}
    \vspace{-12pt}
\end{table*}

\subsection{Main Results and Comparison}
\noindent\textbf{NuScenes dataset.} To widely compare with existing end-to-end planning methods in the community, we evaluate \ourmethod{} on nuScenes dataset~\cite{caesar2020nuscenes}. Tab.~\ref{tab:nuscenes} show that \ourmethod{} notably outperforms all the previous algorithms despite its self-supervised training format. Unlike existing works, \ourmethod{} does not require any perception pretraining or human annotations. This self-supervision nature enables \ourmethod{} to utilize all the accessible raw trajectory data, whereas these easily collectable raw driving logs offer limited benefit to other algorithms without a labeling process.

\noindent\textbf{\ourdataset{} benchmark.} 
 In Tab.~\ref{tab:womd}, we primarily compare \ourmethod{} with the vanilla PaLI3-5B baseline (Sec.~\ref{sec:vanilla}) and modular algorithm MotionLM~\cite{seff2023motionlm}. Compared with vanilla PaLI3-5B, a notable gap is witnessed for both sample-wise and behavior-wise metrics. For reference, we also adapt a latest motion prediction algorithms MotionLM~\cite{seff2023motionlm} (enhanced internally reproduced version) for planning task by only predicting the future trajectories of ego-vehicle and injecting the high-level command to the model. Because it utilizes auto-labelled high-quality objects, tracklets, and roadgraph signals as model inputs~\cite{Qi2021Offboard3O}, a direct comparison with our end-to-end approaches is not equitable.
However, as shown in Tab.~\ref{tab:womd}, \ourmethod{} still achieves favorable performance compared to MotionLM, particularly in behavior-wise metrics even if \ourmethod{} uses only raw camera images as inputs.

\subsection{Analysis}
\begin{figure}[t]
    \centering
    \begin{subfigure}{1.0\linewidth}
        \centering
        \includegraphics[width=0.32\linewidth]{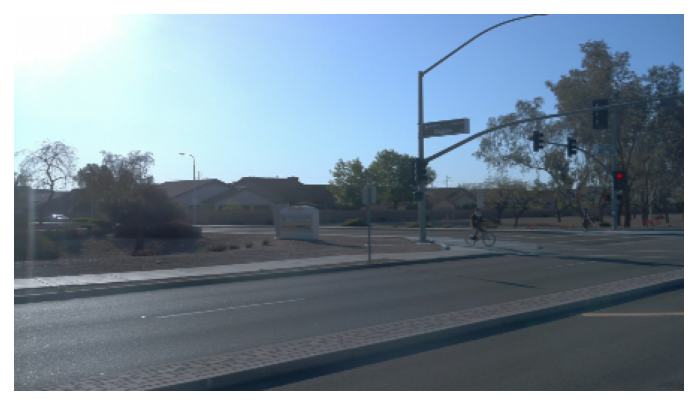}
        \includegraphics[width=0.32\linewidth]{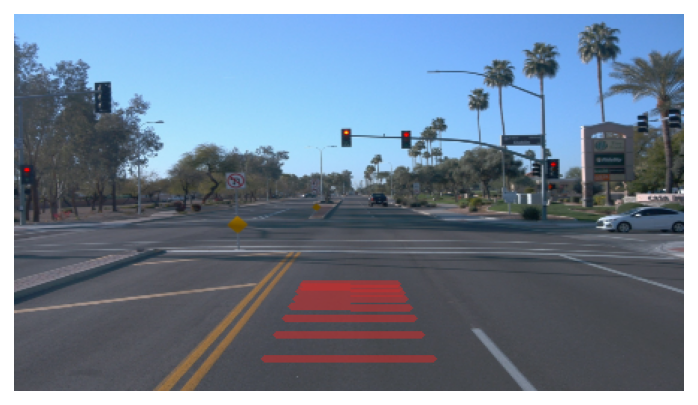}
        \includegraphics[width=0.32\linewidth]{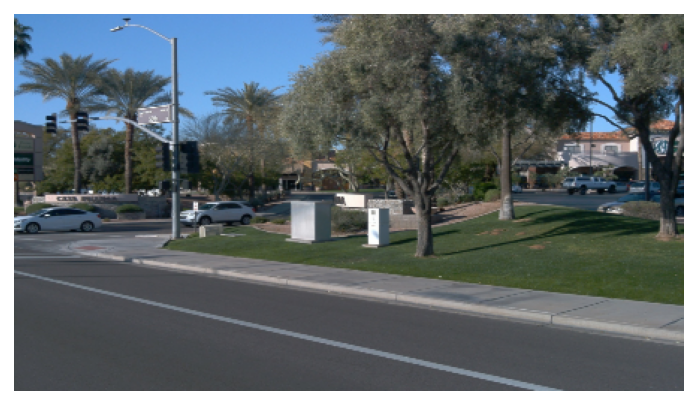}
        \vspace{-5pt}
        \caption{The ego-vehicle decelerates to wait for the red lights.}
    \end{subfigure}
    \begin{subfigure}{1.0\linewidth}
        \centering
        \includegraphics[width=0.32\linewidth]{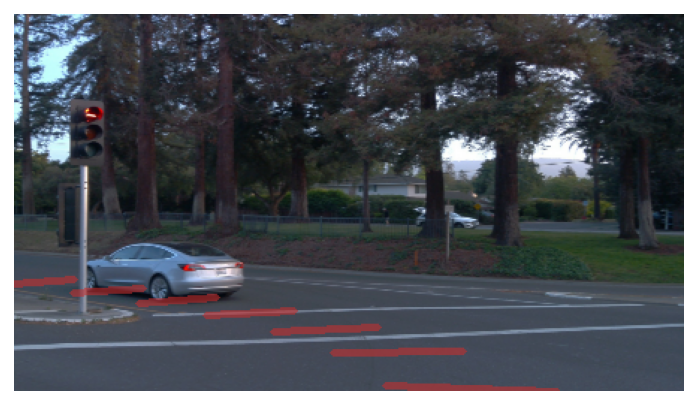}
        \includegraphics[width=0.32\linewidth]{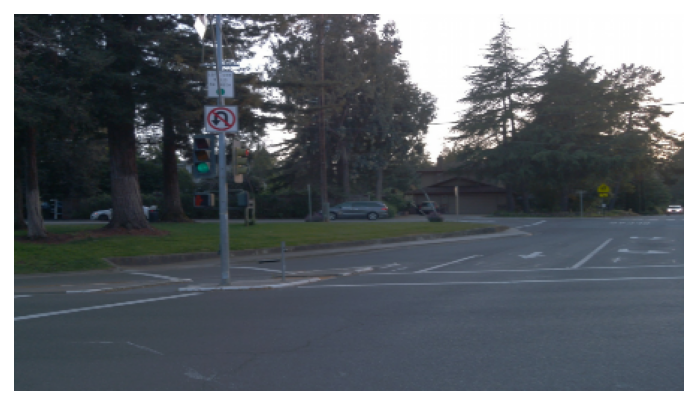}
        \includegraphics[width=0.32\linewidth]{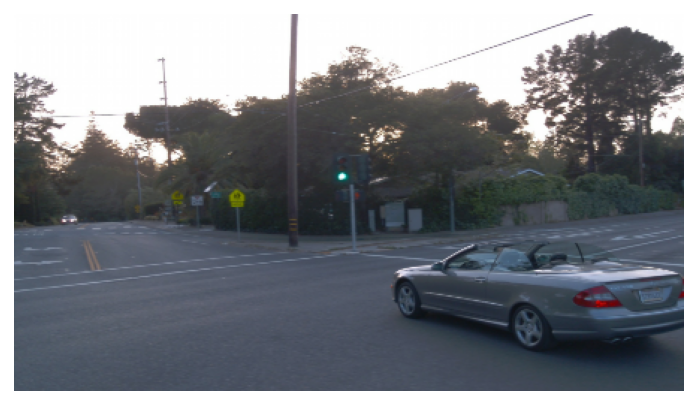}
        \vspace{-5pt}
        \caption{The ego-vehicle conducts a U-turn at the intersection.}
    \end{subfigure}
    \begin{subfigure}{1.0\linewidth}
        \centering
        \includegraphics[width=0.32\linewidth]{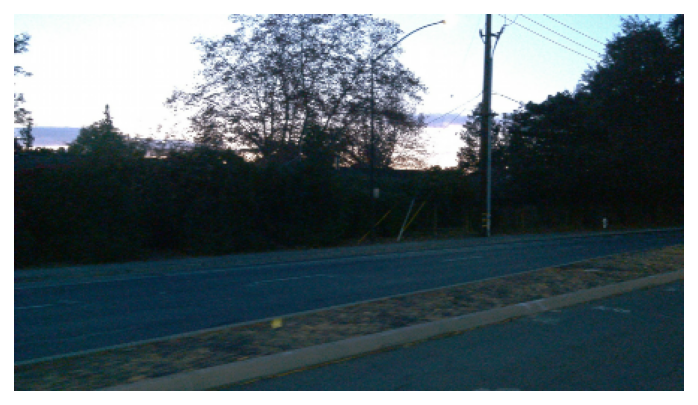}
        \includegraphics[width=0.32\linewidth]{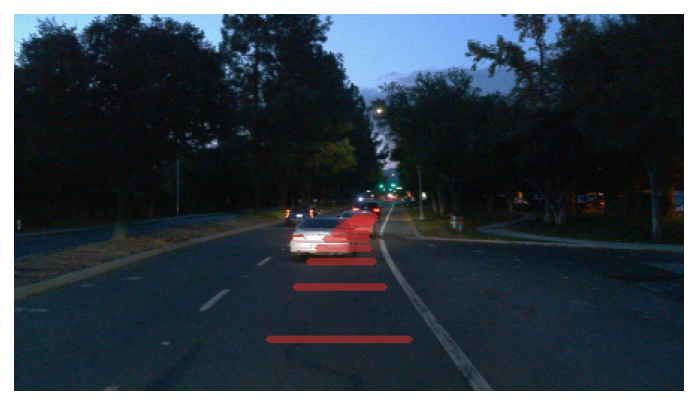}
        \includegraphics[width=0.32\linewidth]{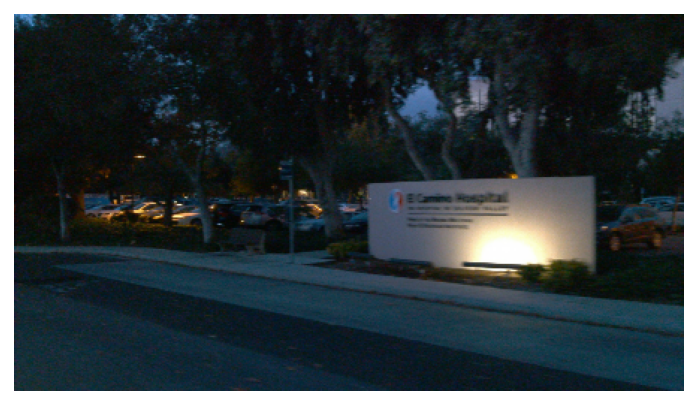}
        \vspace{-5pt}
        \caption{The ego-vehicle keeps the lane under bad lighting condition.}
    \end{subfigure}
    \vspace{-10pt}
    \caption{Qualitative results of motion planning. We show the \textit{front left, front, front right} cameras for each case.}
    \label{fig:visualize}
    \vspace{-10pt}
\end{figure}

\textbf{Qualitative results.}
Fig.~\ref{fig:visualize} visualizes planning results in diverse scenarios. Our proposed \ourmethod{} can determine the future ego-behavior based on the traffic lights and road lane. It can also complete challenging behaviors and deal with different lighting conditions. We will include more visualizations in supplementary materials.

\begin{figure}[ht]
    \centering
    \begin{minipage}{0.48\linewidth}
        \centering
        \includegraphics[width=\linewidth]{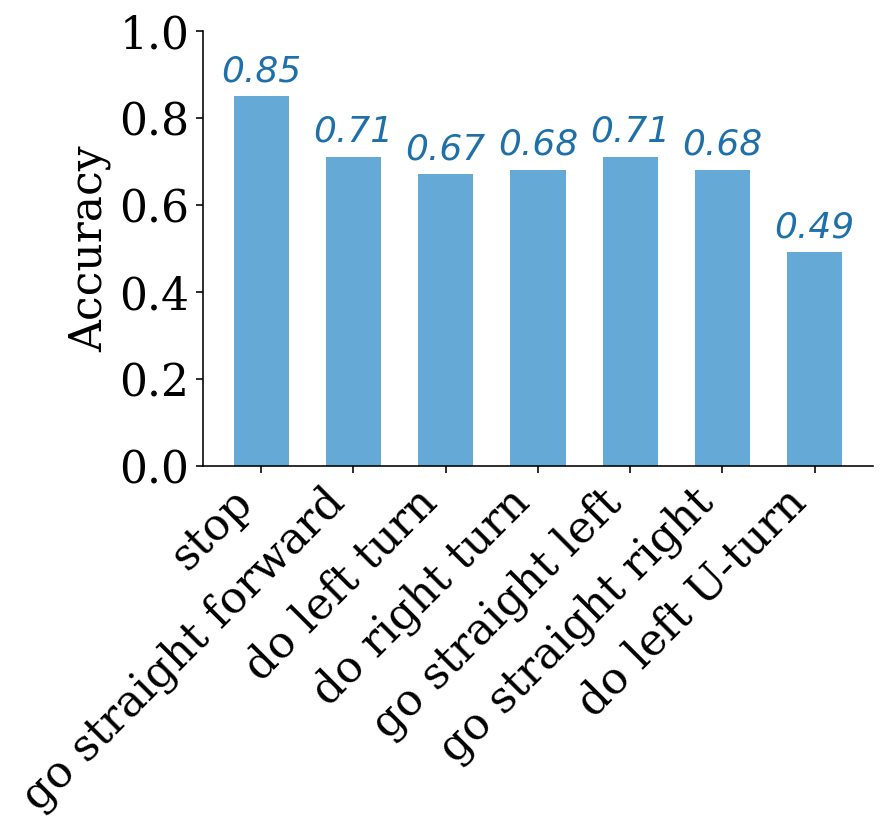}
        \vspace{-20pt}
        \caption{Meta-decision prediction accuracy for each behavior.}
        \label{fig:decision}
    \end{minipage}
    \hfill
    \begin{minipage}{0.48\linewidth}
        \centering
        \includegraphics[width=\linewidth]{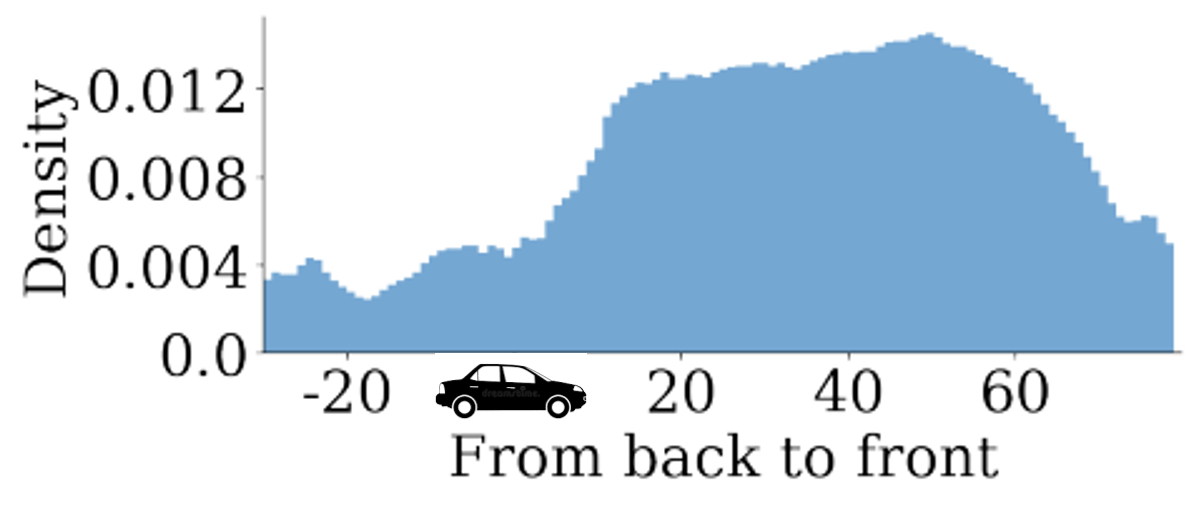}
        \includegraphics[width=\linewidth]{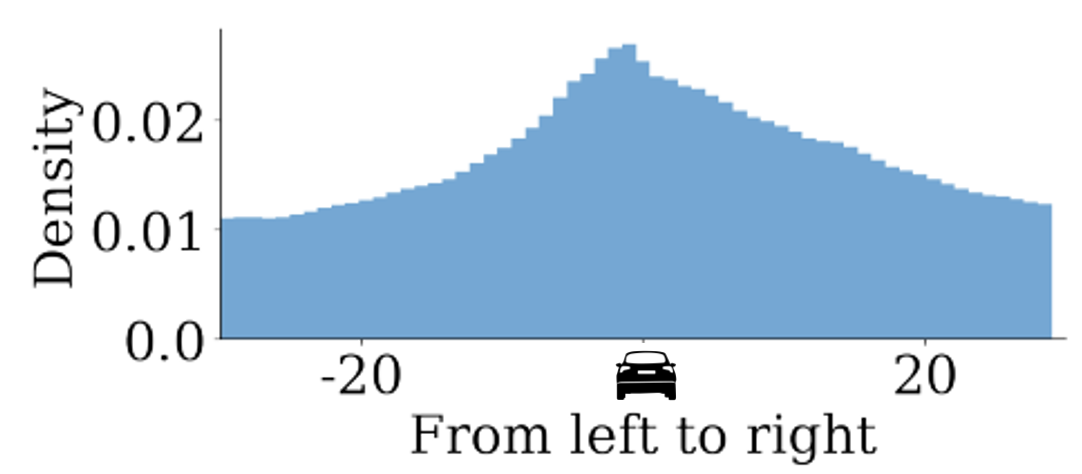}
        \caption{Distribution of sparse volumes along x- and y-axes.}
        \label{fig:sparse}
    \end{minipage}
    \vspace{-15pt}
\end{figure}

\noindent\textbf{Meta-decision reliability.} Fig.~\ref{fig:decision} demonstrates the accuracy of meta-decision prediction (Sec.~\ref{sec:decision}) on \ourdataset{} validation set. Across all behaviors, the model provides reliable meta-decision estimations. Without any human annotations, this preliminary prediction can simplify the derivation of numeric motion planning.

\noindent\textbf{Distribution of sparse volumes.} We visualize the distribution of self-supervised learned sparse volumes along x-axis and y-axis over \ourdataset{} validation set in Fig.~\ref{fig:sparse}. From back to front, the sparse volumes concentrate on the close front region. From left to right, the sparse volumes cover all the regions, since there are turning scenarios, but most volumes focus on the middle region. These distributions are consistent with human driving experiences.

\subsection{Ablation Studies}
This section presents experiments conducted on the WOMD planning benchmark to clarify the roles of specific modules within our \ourmethod{} algorithm. More ablation studies are available in the supplementary material. 

\noindent\textbf{MLLM inputs.} In Tab.~\ref{tab:ablation-inputs}, we analyze the roles of camera images and historical ego-states separately using the vanilla PaLI model. Consistent with \cite{zhai2023ADMLP,li2024ego}, the importance of ego-states is acknowledged. Meanwhile, different from their observations on nuScenes benchmark, camera inputs also make great differences on \ourdataset{}. We assume that \ourdataset{} covers more diverse driving scenarios including many severe changes of speed and directions, which attaches importance to the sensor data. This also showcases the advantages of \ourdataset{} for comprehensive evaluation. Tab.~\ref{tab:ablation-inputs} also shows that random initialized model cannot converge without MLLM pretraining. Despite different domains, \ourmethod{} can profit from the strong reasoning ability from large-scale MLLM pretraining on general tasks.

\noindent\textbf{MLLM capability.} In addition to the PaLI3-5B~\cite{chen2023pali} used in other parts, we also tailor PaLI2-3B~\cite{chenpali} for motion planning with the same techniques. PaLI2-3B is weaker in parameter numbers (3B \vs 5B), architecture, pretraining manner, and input resolution (224 \vs 448). As in Tab.~\ref{tab:ablation-model}, \ourmethod{} based on PaLI2-3B demonstrates much inferior performance compared to that based on PaLI3-5B. We conduct experiments on two different training data scales on \ourdataset{}, \textit{i.e.} 20k (nuScenes scale) \vs 400k (full \ourdataset{}). The gap is particularly large given enough training data. It also justifies the necessity to conduct experiments on large-scale dataset, which can fully unleash the potential of powerful MLLMs.

\begin{table}[t!]
    \centering
    \resizebox{\linewidth}{!}{
    \begin{tabular}{c|cc|cc}
    \toprule
      \multirow{2}{*}{\textbf{MLLM}}& \multicolumn{2}{c|}{\ourdataset{} Subset (20k)} &  \multicolumn{2}{c}{Full \ourdataset{} (400k)} \\
        & ADE@5s & bADE@5s &  ADE@5s & bADE@5s\\
    \midrule
        PaLI2-3B~\cite{chenpali} & 1.241 & 1.846 & 1.035 & 1.522 \\
        PaLI3-5B~\cite{chen2023pali} & \textbf{1.075} & \textbf{1.495} & \textbf{0.693} & \textbf{0.928} \\
    \bottomrule
    \end{tabular}
    }
    \vspace{-10pt}
    \caption{Ablation studies on planning performance scalabity of our methods \wrt model and data sizes. We experiment with two MLLMs on two different training data sizes.}
    \label{tab:ablation-model}
\end{table}

\noindent\textbf{Sparse volume resolution.} Tab.~\ref{tab:ablation-resolution} shows the results with different resolution for sparse volumes with same number of sparse volumes ($M=6000$). In line with Fig.~\ref{fig:roadmap}, low resolution leads to relatively worse performance since it limits the precision of 3D spatial reasoning. Interestingly, higher resolution along $z$-axis does not definitely improve the model performance since the motion planning mainly works on the $xy$-plane and too low sparsity ratio tends to make the optimization unstable.

\begin{table}[t]
    \centering
    \begin{tabular}{c|cc}
       \toprule
       \textbf{Method} & \textbf{ADE@5s} & \textbf{bADE@5s}\\
       \midrule
        vanilla PaLI & \textbf{0.798} & \textbf{1.069}  \\
        \midrule
        \textit{w/o historical states} & 2.009 & 2.250\\
        \textit{w/o camera images} & 1.081 & 1.448 \\
        \midrule
        \textit{w/o MLLM pretraining} & \multicolumn{2}{c}{\textit{fail to converge}}\\
    \bottomrule
    \end{tabular}
    \vspace{-10pt}
    \caption{Ablation studies on MLLM inputs.}
    \label{tab:ablation-inputs}
\end{table}

\begin{table}[t]
    \centering
    \begin{tabular}{c|cc}
       \toprule
        \textbf{Resolution} & \textbf{ADE@5s} & \textbf{bADE@5s}\\
       \midrule
        $2m\times 2m\times 2m$ & 0.798 & 1.201  \\
        $1m\times 1m\times 2m$ & \textbf{0.750} &\textbf{1.005}\\
        $1m\times 1m\times 1m$ & 0.765 & 1.051 \\
    \bottomrule
    \end{tabular}
    \vspace{-10pt}
    \caption{Ablation studies on sparse volume resolutions.}
    \label{tab:ablation-resolution}
    \vspace{-10pt}
\end{table}

\section{Conclusion and Future Work}
This paper introduces \ourmethod{}, a scalable self-supervised motion planning framework for autonomous driving that utilizes Multimodal Large Language Models (MLLMs). To enhance 3D reasoning in MLLMs, we propose a novel sparse volume representation, which enables improved spatio-temporal reasoning through the aggregation of multi-view and multi-frame image inputs. Additionally, we design behavior-wise metrics for comprehensive evaluation on the large-scale \ourdataset{} benchmark. \ourmethod{} achieves state-of-the-art performance on both nuScenes and \ourdataset{} benchmarks, without requiring any human annotations. This demonstrates the potential of self-supervised learning for end-to-end autonomous driving. Future work will involve applying our methods to other powerful MLLM architectures. Our contributions are orthogonal to previous works, such as those that explore multi-task learning or CoT reasoning~\cite{simadrivelm,wang2024omnidrive}. Combining our large-scale self-supervised learning approach with supervised finetuning on targeted small-scale labeled data may further enhance the final performance and interpretability of the system. 


{
    \small
    \bibliographystyle{ieeenat_fullname}
    \bibliography{main}

\begin{thebibliography}{66}
\providecommand{\natexlab}[1]{#1}
\providecommand{\url}[1]{\texttt{#1}}
\expandafter\ifx\csname urlstyle\endcsname\relax
  \providecommand{\doi}[1]{doi: #1}\else
  \providecommand{\doi}{doi: \begingroup \urlstyle{rm}\Url}\fi

\bibitem[Achiam et~al.(2023)Achiam, Adler, Agarwal, Ahmad, Akkaya, Aleman,
  Almeida, Altenschmidt, Altman, Anadkat, et~al.]{achiam2023gpt}
Josh Achiam, Steven Adler, Sandhini Agarwal, Lama Ahmad, Ilge Akkaya,
  Florencia~Leoni Aleman, Diogo Almeida, Janko Altenschmidt, Sam Altman,
  Shyamal Anadkat, et~al.
\newblock Gpt-4 technical report.
\newblock \emph{arXiv preprint arXiv:2303.08774}, 2023.

\bibitem[Alayrac et~al.(2022)Alayrac, Donahue, Luc, Miech, Barr, Hasson, Lenc,
  Mensch, Millican, Reynolds, et~al.]{alayrac2022flamingo}
Jean-Baptiste Alayrac, Jeff Donahue, Pauline Luc, Antoine Miech, Iain Barr,
  Yana Hasson, Karel Lenc, Arthur Mensch, Katherine Millican, Malcolm Reynolds,
  et~al.
\newblock Flamingo: a visual language model for few-shot learning.
\newblock \emph{Advances in neural information processing systems},
  35:\penalty0 23716--23736, 2022.

\bibitem[Bai et~al.(2023)Bai, Bai, Yang, Wang, Tan, Wang, Lin, Zhou, and
  Zhou]{bai2023qwen}
Jinze Bai, Shuai Bai, Shusheng Yang, Shijie Wang, Sinan Tan, Peng Wang, Junyang
  Lin, Chang Zhou, and Jingren Zhou.
\newblock Qwen-vl: A versatile vision-language model for understanding,
  localization, text reading, and beyond.
\newblock \emph{arXiv preprint arXiv:2308.12966}, 1\penalty0 (2):\penalty0 3,
  2023.

\bibitem[Beyer et~al.(2024)Beyer, Steiner, Pinto, Kolesnikov, Wang, Salz,
  Neumann, Alabdulmohsin, Tschannen, Bugliarello, et~al.]{beyer2024paligemma}
Lucas Beyer, Andreas Steiner, Andr{\'e}~Susano Pinto, Alexander Kolesnikov,
  Xiao Wang, Daniel Salz, Maxim Neumann, Ibrahim Alabdulmohsin, Michael
  Tschannen, Emanuele Bugliarello, et~al.
\newblock Paligemma: A versatile 3b vlm for transfer.
\newblock \emph{arXiv preprint arXiv:2407.07726}, 2024.

\bibitem[Caesar et~al.(2020)Caesar, Bankiti, Lang, Vora, Liong, Xu, Krishnan,
  Pan, Baldan, and Beijbom]{caesar2020nuscenes}
Holger Caesar, Varun Bankiti, Alex~H Lang, Sourabh Vora, Venice~Erin Liong,
  Qiang Xu, Anush Krishnan, Yu Pan, Giancarlo Baldan, and Oscar Beijbom.
\newblock nuscenes: A multimodal dataset for autonomous driving.
\newblock In \emph{Proceedings of the IEEE/CVF conference on computer vision
  and pattern recognition}, pages 11621--11631, 2020.

\bibitem[Casas et~al.(2021)Casas, Sadat, and Urtasun]{casas2021mp3}
Sergio Casas, Abbas Sadat, and Raquel Urtasun.
\newblock Mp3: A unified model to map, perceive, predict and plan.
\newblock In \emph{Proceedings of the IEEE/CVF Conference on Computer Vision
  and Pattern Recognition}, pages 14403--14412, 2021.

\bibitem[Chen et~al.(2024{\natexlab{a}})Chen, Xu, Kirmani, Ichter, Sadigh,
  Guibas, and Xia]{chen2024spatialvlm}
Boyuan Chen, Zhuo Xu, Sean Kirmani, Brain Ichter, Dorsa Sadigh, Leonidas
  Guibas, and Fei Xia.
\newblock Spatialvlm: Endowing vision-language models with spatial reasoning
  capabilities.
\newblock In \emph{Proceedings of the IEEE/CVF Conference on Computer Vision
  and Pattern Recognition}, pages 14455--14465, 2024{\natexlab{a}}.

\bibitem[Chen et~al.(2024{\natexlab{b}})Chen, Sinavski, H{\"u}nermann,
  Karnsund, Willmott, Birch, Maund, and Shotton]{chen2024driving}
Long Chen, Oleg Sinavski, Jan H{\"u}nermann, Alice Karnsund, Andrew~James
  Willmott, Danny Birch, Daniel Maund, and Jamie Shotton.
\newblock Driving with llms: Fusing object-level vector modality for
  explainable autonomous driving.
\newblock In \emph{2024 IEEE International Conference on Robotics and
  Automation (ICRA)}, pages 14093--14100. IEEE, 2024{\natexlab{b}}.

\bibitem[Chen et~al.()Chen, Wang, Changpinyo, Piergiovanni, Padlewski, Salz,
  Goodman, Grycner, Mustafa, Beyer, et~al.]{chenpali}
Xi Chen, Xiao Wang, Soravit Changpinyo, AJ Piergiovanni, Piotr Padlewski,
  Daniel Salz, Sebastian Goodman, Adam Grycner, Basil Mustafa, Lucas Beyer,
  et~al.
\newblock Pali: A jointly-scaled multilingual language-image model.
\newblock In \emph{The Eleventh International Conference on Learning
  Representations}.

\bibitem[Chen et~al.(2023)Chen, Wang, Beyer, Kolesnikov, Wu, Voigtlaender,
  Mustafa, Goodman, Alabdulmohsin, Padlewski, et~al.]{chen2023pali}
Xi Chen, Xiao Wang, Lucas Beyer, Alexander Kolesnikov, Jialin Wu, Paul
  Voigtlaender, Basil Mustafa, Sebastian Goodman, Ibrahim Alabdulmohsin, Piotr
  Padlewski, et~al.
\newblock Pali-3 vision language models: Smaller, faster, stronger.
\newblock \emph{arXiv preprint arXiv:2310.09199}, 2023.

\bibitem[Cheng et~al.(2024)Cheng, Yin, Fu, Guo, Yang, Kautz, Wang, and
  Liu]{cheng2024spatialrgpt}
An-Chieh Cheng, Hongxu Yin, Yang Fu, Qiushan Guo, Ruihan Yang, Jan Kautz,
  Xiaolong Wang, and Sifei Liu.
\newblock Spatialrgpt: Grounded spatial reasoning in vision language model.
\newblock \emph{arXiv preprint arXiv:2406.01584}, 2024.

\bibitem[Chung et~al.(2024)Chung, Hou, Longpre, Zoph, Tay, Fedus, Li, Wang,
  Dehghani, Brahma, et~al.]{chung2024scaling}
Hyung~Won Chung, Le Hou, Shayne Longpre, Barret Zoph, Yi Tay, William Fedus,
  Yunxuan Li, Xuezhi Wang, Mostafa Dehghani, Siddhartha Brahma, et~al.
\newblock Scaling instruction-finetuned language models.
\newblock \emph{Journal of Machine Learning Research}, 25\penalty0
  (70):\penalty0 1--53, 2024.

\bibitem[Codevilla et~al.(2018)Codevilla, M{\"u}ller, L{\'o}pez, Koltun, and
  Dosovitskiy]{codevilla2018end}
Felipe Codevilla, Matthias M{\"u}ller, Antonio L{\'o}pez, Vladlen Koltun, and
  Alexey Dosovitskiy.
\newblock End-to-end driving via conditional imitation learning.
\newblock In \emph{2018 IEEE international conference on robotics and
  automation (ICRA)}, pages 4693--4700. IEEE, 2018.

\bibitem[Dai et~al.(2023)Dai, Li, Li, Tiong, Zhao, Wang, Li, Fung, and
  Hoi]{instructblip}
Wenliang Dai, Junnan Li, Dongxu Li, Anthony Meng~Huat Tiong, Junqi Zhao,
  Weisheng Wang, Boyang Li, Pascale Fung, and Steven Hoi.
\newblock Instructblip: Towards general-purpose vision-language models with
  instruction tuning, 2023.

\bibitem[Ding et~al.(2024)Ding, Han, Xu, Liang, Zhang, and
  Li]{ding2024holistic}
Xinpeng Ding, Jianhua Han, Hang Xu, Xiaodan Liang, Wei Zhang, and Xiaomeng Li.
\newblock Holistic autonomous driving understanding by bird's-eye-view injected
  multi-modal large models.
\newblock In \emph{Proceedings of the IEEE/CVF Conference on Computer Vision
  and Pattern Recognition}, pages 13668--13677, 2024.

\bibitem[Dosovitskiy et~al.(2017)Dosovitskiy, Ros, Codevilla, Lopez, and
  Koltun]{dosovitskiy2017carla}
Alexey Dosovitskiy, German Ros, Felipe Codevilla, Antonio Lopez, and Vladlen
  Koltun.
\newblock Carla: An open urban driving simulator.
\newblock In \emph{Conference on robot learning}, pages 1--16. PMLR, 2017.

\bibitem[Dosovitskiy et~al.(2020)Dosovitskiy, Beyer, Kolesnikov, Weissenborn,
  Zhai, Unterthiner, Dehghani, Minderer, Heigold, Gelly,
  et~al.]{dosovitskiy2020image}
Alexey Dosovitskiy, Lucas Beyer, Alexander Kolesnikov, Dirk Weissenborn,
  Xiaohua Zhai, Thomas Unterthiner, Mostafa Dehghani, Matthias Minderer, Georg
  Heigold, Sylvain Gelly, et~al.
\newblock An image is worth 16x16 words: Transformers for image recognition at
  scale.
\newblock \emph{arXiv preprint arXiv:2010.11929}, 2020.

\bibitem[Ettinger et~al.(2021)Ettinger, Cheng, Caine, Liu, Zhao, Pradhan, Chai,
  Sapp, Qi, Zhou, et~al.]{ettinger2021large}
Scott Ettinger, Shuyang Cheng, Benjamin Caine, Chenxi Liu, Hang Zhao, Sabeek
  Pradhan, Yuning Chai, Ben Sapp, Charles~R Qi, Yin Zhou, et~al.
\newblock Large scale interactive motion forecasting for autonomous driving:
  The waymo open motion dataset.
\newblock In \emph{Proceedings of the IEEE/CVF International Conference on
  Computer Vision}, pages 9710--9719, 2021.

\bibitem[Fang et~al.(2023)Fang, Wang, Xie, Sun, Wu, Wang, Huang, Wang, and
  Cao]{fang2023eva}
Yuxin Fang, Wen Wang, Binhui Xie, Quan Sun, Ledell Wu, Xinggang Wang, Tiejun
  Huang, Xinlong Wang, and Yue Cao.
\newblock Eva: Exploring the limits of masked visual representation learning at
  scale.
\newblock In \emph{Proceedings of the IEEE/CVF Conference on Computer Vision
  and Pattern Recognition}, pages 19358--19369, 2023.

\bibitem[Harley et~al.(2023)Harley, Fang, Li, Ambrus, and
  Fragkiadaki]{harley2023simple}
Adam~W Harley, Zhaoyuan Fang, Jie Li, Rares Ambrus, and Katerina Fragkiadaki.
\newblock Simple-bev: What really matters for multi-sensor bev perception?
\newblock In \emph{2023 IEEE International Conference on Robotics and
  Automation (ICRA)}, pages 2759--2765. IEEE, 2023.

\bibitem[Holtzman et~al.(2019)Holtzman, Buys, Du, Forbes, and
  Choi]{holtzman2019curious}
Ari Holtzman, Jan Buys, Li Du, Maxwell Forbes, and Yejin Choi.
\newblock The curious case of neural text degeneration.
\newblock \emph{arXiv preprint arXiv:1904.09751}, 2019.

\bibitem[Hong et~al.(2023)Hong, Zhen, Chen, Zheng, Du, Chen, and
  Gan]{hong20233d}
Yining Hong, Haoyu Zhen, Peihao Chen, Shuhong Zheng, Yilun Du, Zhenfang Chen,
  and Chuang Gan.
\newblock 3d-llm: Injecting the 3d world into large language models.
\newblock \emph{Advances in Neural Information Processing Systems},
  36:\penalty0 20482--20494, 2023.

\bibitem[Hu et~al.(2021)Hu, Shen, Wallis, Allen-Zhu, Li, Wang, Wang, and
  Chen]{hu2021lora}
Edward~J Hu, Yelong Shen, Phillip Wallis, Zeyuan Allen-Zhu, Yuanzhi Li, Shean
  Wang, Lu Wang, and Weizhu Chen.
\newblock Lora: Low-rank adaptation of large language models.
\newblock \emph{arXiv preprint arXiv:2106.09685}, 2021.

\bibitem[Hu et~al.(2022)Hu, Chen, Wu, Li, Yan, and Tao]{hu2022st}
Shengchao Hu, Li Chen, Penghao Wu, Hongyang Li, Junchi Yan, and Dacheng Tao.
\newblock St-p3: End-to-end vision-based autonomous driving via
  spatial-temporal feature learning.
\newblock In \emph{European Conference on Computer Vision}, pages 533--549.
  Springer, 2022.

\bibitem[Hu et~al.(2023)Hu, Yang, Chen, Li, Sima, Zhu, Chai, Du, Lin, Wang,
  et~al.]{hu2023planning}
Yihan Hu, Jiazhi Yang, Li Chen, Keyu Li, Chonghao Sima, Xizhou Zhu, Siqi Chai,
  Senyao Du, Tianwei Lin, Wenhai Wang, et~al.
\newblock Planning-oriented autonomous driving.
\newblock In \emph{Proceedings of the IEEE/CVF Conference on Computer Vision
  and Pattern Recognition}, pages 17853--17862, 2023.

\bibitem[Hwang et~al.(2024)Hwang, Xu, Lin, Hung, Ji, Choi, Huang, He,
  Covington, Sapp, Zhou, Guo, Anguelov, and
  Tan]{hwang2024emmaendtoendmultimodalmodel}
Jyh-Jing Hwang, Runsheng Xu, Hubert Lin, Wei-Chih Hung, Jingwei Ji, Kristy
  Choi, Di Huang, Tong He, Paul Covington, Benjamin Sapp, Yin Zhou, James Guo,
  Dragomir Anguelov, and Mingxing Tan.
\newblock Emma: End-to-end multimodal model for autonomous driving, 2024.

\bibitem[Igl et~al.(2022)Igl, Kim, Kuefler, Mougin, Shah, Shiarlis, Anguelov,
  Palatucci, White, and Whiteson]{igl2022symphony}
Maximilian Igl, Daewoo Kim, Alex Kuefler, Paul Mougin, Punit Shah, Kyriacos
  Shiarlis, Dragomir Anguelov, Mark Palatucci, Brandyn White, and Shimon
  Whiteson.
\newblock Symphony: Learning realistic and diverse agents for autonomous
  driving simulation.
\newblock In \emph{2022 International Conference on Robotics and Automation
  (ICRA)}, pages 2445--2451. IEEE, 2022.

\bibitem[Jiang et~al.(2023)Jiang, Chen, Xu, Liao, Chen, Zhou, Zhang, Liu,
  Huang, and Wang]{jiang2023vad}
Bo Jiang, Shaoyu Chen, Qing Xu, Bencheng Liao, Jiajie Chen, Helong Zhou, Qian
  Zhang, Wenyu Liu, Chang Huang, and Xinggang Wang.
\newblock Vad: Vectorized scene representation for efficient autonomous
  driving.
\newblock In \emph{Proceedings of the IEEE/CVF International Conference on
  Computer Vision}, pages 8340--8350, 2023.

\bibitem[Kim et~al.(2018)Kim, Rohrbach, Darrell, Canny, and
  Akata]{kim2018textual}
Jinkyu Kim, Anna Rohrbach, Trevor Darrell, John Canny, and Zeynep Akata.
\newblock Textual explanations for self-driving vehicles.
\newblock In \emph{Proceedings of the European conference on computer vision
  (ECCV)}, pages 563--578, 2018.

\bibitem[Li et~al.(2023)Li, Li, Savarese, and Hoi]{li2023blip}
Junnan Li, Dongxu Li, Silvio Savarese, and Steven Hoi.
\newblock Blip-2: Bootstrapping language-image pre-training with frozen image
  encoders and large language models.
\newblock In \emph{International conference on machine learning}, pages
  19730--19742. PMLR, 2023.

\bibitem[Li et~al.(2022)Li, Wang, Li, Xie, Sima, Lu, Qiao, and
  Dai]{li2022bevformer}
Zhiqi Li, Wenhai Wang, Hongyang Li, Enze Xie, Chonghao Sima, Tong Lu, Yu Qiao,
  and Jifeng Dai.
\newblock Bevformer: Learning bird’s-eye-view representation from
  multi-camera images via spatiotemporal transformers.
\newblock In \emph{European conference on computer vision}, pages 1--18.
  Springer, 2022.

\bibitem[Li et~al.(2024)Li, Yu, Lan, Li, Kautz, Lu, and Alvarez]{li2024ego}
Zhiqi Li, Zhiding Yu, Shiyi Lan, Jiahan Li, Jan Kautz, Tong Lu, and Jose~M
  Alvarez.
\newblock Is ego status all you need for open-loop end-to-end autonomous
  driving?
\newblock In \emph{Proceedings of the IEEE/CVF Conference on Computer Vision
  and Pattern Recognition}, pages 14864--14873, 2024.

\bibitem[Liao et~al.()Liao, Chen, Wang, Cheng, Zhang, Liu, and
  Huang]{liaomaptr}
Bencheng Liao, Shaoyu Chen, Xinggang Wang, Tianheng Cheng, Qian Zhang, Wenyu
  Liu, and Chang Huang.
\newblock Maptr: Structured modeling and learning for online vectorized hd map
  construction.
\newblock In \emph{The Eleventh International Conference on Learning
  Representations}.

\bibitem[Liu et~al.(2024)Liu, Li, Wu, and Lee]{liu2024visual}
Haotian Liu, Chunyuan Li, Qingyang Wu, and Yong~Jae Lee.
\newblock Visual instruction tuning.
\newblock \emph{Advances in neural information processing systems}, 36, 2024.

\bibitem[Liu et~al.(2023)Liu, Tang, Amini, Yang, Mao, Rus, and
  Han]{liu2023bevfusion}
Zhijian Liu, Haotian Tang, Alexander Amini, Xinyu Yang, Huizi Mao, Daniela~L
  Rus, and Song Han.
\newblock Bevfusion: Multi-task multi-sensor fusion with unified bird's-eye
  view representation.
\newblock In \emph{2023 IEEE international conference on robotics and
  automation (ICRA)}, pages 2774--2781. IEEE, 2023.

\bibitem[Ma et~al.(2024)Ma, Bhalgat, Smart, Chen, Li, Ding, Gu, Chen, Peng,
  Bian, et~al.]{ma2024llms}
Xianzheng Ma, Yash Bhalgat, Brandon Smart, Shuai Chen, Xinghui Li, Jian Ding,
  Jindong Gu, Dave~Zhenyu Chen, Songyou Peng, Jia-Wang Bian, et~al.
\newblock When llms step into the 3d world: A survey and meta-analysis of 3d
  tasks via multi-modal large language models.
\newblock \emph{arXiv preprint arXiv:2405.10255}, 2024.

\bibitem[Ma et~al.(2023)Ma, Cao, Sun, Pavone, and Xiao]{ma2023dolphins}
Yingzi Ma, Yulong Cao, Jiachen Sun, Marco Pavone, and Chaowei Xiao.
\newblock Dolphins: Multimodal language model for driving.
\newblock \emph{arXiv preprint arXiv:2312.00438}, 2023.

\bibitem[Mao et~al.(2023{\natexlab{a}})Mao, Qian, Ye, Zhao, and
  Wang]{mao2023gpt}
Jiageng Mao, Yuxi Qian, Junjie Ye, Hang Zhao, and Yue Wang.
\newblock Gpt-driver: Learning to drive with gpt.
\newblock \emph{arXiv preprint arXiv:2310.01415}, 2023{\natexlab{a}}.

\bibitem[Mao et~al.(2023{\natexlab{b}})Mao, Ye, Qian, Pavone, and
  Wang]{mao2023language}
Jiageng Mao, Junjie Ye, Yuxi Qian, Marco Pavone, and Yue Wang.
\newblock A language agent for autonomous driving.
\newblock \emph{arXiv preprint arXiv:2311.10813}, 2023{\natexlab{b}}.

\bibitem[McKinzie et~al.(2024)McKinzie, Gan, Fauconnier, Dodge, Zhang, Dufter,
  Shah, Du, Peng, Weers, et~al.]{mckinzie2024mm1}
Brandon McKinzie, Zhe Gan, Jean-Philippe Fauconnier, Sam Dodge, Bowen Zhang,
  Philipp Dufter, Dhruti Shah, Xianzhi Du, Futang Peng, Floris Weers, et~al.
\newblock Mm1: Methods, analysis \& insights from multimodal llm pre-training.
\newblock \emph{arXiv preprint arXiv:2403.09611}, 2024.

\bibitem[Nayakanti et~al.(2023)Nayakanti, Al-Rfou, Zhou, Goel, Refaat, and
  Sapp]{nayakanti2023wayformer}
Nigamaa Nayakanti, Rami Al-Rfou, Aurick Zhou, Kratarth Goel, Khaled~S Refaat,
  and Benjamin Sapp.
\newblock Wayformer: Motion forecasting via simple \& efficient attention
  networks.
\newblock In \emph{2023 IEEE International Conference on Robotics and
  Automation (ICRA)}, pages 2980--2987. IEEE, 2023.

\bibitem[OpenAI(2023)]{gpt4v}
OpenAI.
\newblock Gpt-4v(ision) system card, 2023.

\bibitem[Pomerleau(1988)]{pomerleau1988alvinn}
Dean~A Pomerleau.
\newblock Alvinn: An autonomous land vehicle in a neural network.
\newblock \emph{Advances in neural information processing systems}, 1, 1988.

\bibitem[Qi et~al.(2021)Qi, Zhou, Najibi, Sun, Vo, Deng, and
  Anguelov]{Qi2021Offboard3O}
C. Qi, Yin Zhou, Mahyar Najibi, Pei Sun, Khoa~T. Vo, Boyang Deng, and Dragomir
  Anguelov.
\newblock Offboard 3d object detection from point cloud sequences.
\newblock \emph{2021 IEEE/CVF Conference on Computer Vision and Pattern
  Recognition (CVPR)}, pages 6130--6140, 2021.

\bibitem[Radford et~al.(2021)Radford, Kim, Hallacy, Ramesh, Goh, Agarwal,
  Sastry, Askell, Mishkin, Clark, et~al.]{radford2021learning}
Alec Radford, Jong~Wook Kim, Chris Hallacy, Aditya Ramesh, Gabriel Goh,
  Sandhini Agarwal, Girish Sastry, Amanda Askell, Pamela Mishkin, Jack Clark,
  et~al.
\newblock Learning transferable visual models from natural language
  supervision.
\newblock In \emph{International conference on machine learning}, pages
  8748--8763. PMLR, 2021.

\bibitem[Seff et~al.(2023)Seff, Cera, Chen, Ng, Zhou, Nayakanti, Refaat,
  Al-Rfou, and Sapp]{seff2023motionlm}
Ari Seff, Brian Cera, Dian Chen, Mason Ng, Aurick Zhou, Nigamaa Nayakanti,
  Khaled~S Refaat, Rami Al-Rfou, and Benjamin Sapp.
\newblock Motionlm: Multi-agent motion forecasting as language modeling.
\newblock In \emph{Proceedings of the IEEE/CVF International Conference on
  Computer Vision}, pages 8579--8590, 2023.

\bibitem[Sha et~al.(2023)Sha, Mu, Jiang, Chen, Xu, Luo, Li, Tomizuka, Zhan, and
  Ding]{sha2023languagempc}
Hao Sha, Yao Mu, Yuxuan Jiang, Li Chen, Chenfeng Xu, Ping Luo, Shengbo~Eben Li,
  Masayoshi Tomizuka, Wei Zhan, and Mingyu Ding.
\newblock Languagempc: Large language models as decision makers for autonomous
  driving.
\newblock \emph{arXiv preprint arXiv:2310.03026}, 2023.

\bibitem[Shao et~al.(2024)Shao, Hu, Wang, Song, Waslander, Liu, and
  Li]{shao2024lmdrive}
Hao Shao, Yuxuan Hu, Letian Wang, Guanglu Song, Steven~L Waslander, Yu Liu, and
  Hongsheng Li.
\newblock Lmdrive: Closed-loop end-to-end driving with large language models.
\newblock In \emph{Proceedings of the IEEE/CVF Conference on Computer Vision
  and Pattern Recognition}, pages 15120--15130, 2024.

\bibitem[Sima et~al.()Sima, Renz, Chitta, Chen, Zhang, Xie, Bei{\ss}wenger,
  Luo, Geiger, and Li]{simadrivelm}
Chonghao Sima, Katrin Renz, Kashyap Chitta, Li Chen, Hanxue Zhang, Chengen Xie,
  Jens Bei{\ss}wenger, Ping Luo, Andreas Geiger, and Hongyang Li.
\newblock Drivelm: Driving with graph visual question answering.
\newblock In \emph{First Vision and Language for Autonomous Driving and
  Robotics Workshop}.

\bibitem[Tay et~al.(2022)Tay, Dehghani, Tran, Garcia, Wei, Wang, Chung,
  Shakeri, Bahri, Schuster, et~al.]{tay2022ul2}
Yi Tay, Mostafa Dehghani, Vinh~Q Tran, Xavier Garcia, Jason Wei, Xuezhi Wang,
  Hyung~Won Chung, Siamak Shakeri, Dara Bahri, Tal Schuster, et~al.
\newblock Ul2: Unifying language learning paradigms.
\newblock \emph{arXiv preprint arXiv:2205.05131}, 2022.

\bibitem[Team et~al.(2023)Team, Anil, Borgeaud, Alayrac, Yu, Soricut,
  Schalkwyk, Dai, Hauth, Millican, et~al.]{team2023gemini}
Gemini Team, Rohan Anil, Sebastian Borgeaud, Jean-Baptiste Alayrac, Jiahui Yu,
  Radu Soricut, Johan Schalkwyk, Andrew~M Dai, Anja Hauth, Katie Millican,
  et~al.
\newblock Gemini: a family of highly capable multimodal models.
\newblock \emph{arXiv preprint arXiv:2312.11805}, 2023.

\bibitem[Tian et~al.()Tian, Li, Weng, Chen, Schmerling, Wang, Ivanovic, and
  Pavone]{tiantokenize}
Thomas Tian, Boyi Li, Xinshuo Weng, Yuxiao Chen, Edward Schmerling, Yue Wang,
  Boris Ivanovic, and Marco Pavone.
\newblock Tokenize the world into object-level knowledge to address long-tail
  events in autonomous driving.
\newblock In \emph{Workshop on Language and Robot Learning: Language as an
  Interface}.

\bibitem[Tian et~al.(2024)Tian, Gu, Li, Liu, Hu, Wang, Zhan, Jia, Lang, and
  Zhao]{tian2024drivevlm}
Xiaoyu Tian, Junru Gu, Bailin Li, Yicheng Liu, Chenxu Hu, Yang Wang, Kun Zhan,
  Peng Jia, Xianpeng Lang, and Hang Zhao.
\newblock Drivevlm: The convergence of autonomous driving and large
  vision-language models.
\newblock \emph{arXiv preprint arXiv:2402.12289}, 2024.

\bibitem[Varadarajan et~al.(2022)Varadarajan, Hefny, Srivastava, Refaat,
  Nayakanti, Cornman, Chen, Douillard, Lam, Anguelov,
  et~al.]{varadarajan2022multipath++}
Balakrishnan Varadarajan, Ahmed Hefny, Avikalp Srivastava, Khaled~S Refaat,
  Nigamaa Nayakanti, Andre Cornman, Kan Chen, Bertrand Douillard, Chi~Pang Lam,
  Dragomir Anguelov, et~al.
\newblock Multipath++: Efficient information fusion and trajectory aggregation
  for behavior prediction.
\newblock In \emph{2022 International Conference on Robotics and Automation
  (ICRA)}, pages 7814--7821. IEEE, 2022.

\bibitem[Wang et~al.(2024)Wang, Yu, Jiang, Lan, Shi, Chang, Kautz, Li, and
  Alvarez]{wang2024omnidrive}
Shihao Wang, Zhiding Yu, Xiaohui Jiang, Shiyi Lan, Min Shi, Nadine Chang, Jan
  Kautz, Ying Li, and Jose~M Alvarez.
\newblock Omnidrive: A holistic llm-agent framework for autonomous driving with
  3d perception, reasoning and planning.
\newblock \emph{arXiv preprint arXiv:2405.01533}, 2024.

\bibitem[Wei et~al.(2022)Wei, Wang, Schuurmans, Bosma, Xia, Chi, Le, Zhou,
  et~al.]{wei2022chain}
Jason Wei, Xuezhi Wang, Dale Schuurmans, Maarten Bosma, Fei Xia, Ed Chi, Quoc~V
  Le, Denny Zhou, et~al.
\newblock Chain-of-thought prompting elicits reasoning in large language
  models.
\newblock \emph{Advances in neural information processing systems},
  35:\penalty0 24824--24837, 2022.

\bibitem[Weng et~al.(2024)Weng, Ivanovic, Wang, Wang, and
  Pavone]{weng2024drive}
Xinshuo Weng, Boris Ivanovic, Yan Wang, Yue Wang, and Marco Pavone.
\newblock Para-drive: Parallelized architecture for real-time autonomous
  driving.
\newblock In \emph{Proceedings of the IEEE/CVF Conference on Computer Vision
  and Pattern Recognition}, pages 15449--15458, 2024.

\bibitem[Wu et~al.(2022)Wu, Jia, Chen, Yan, Li, and Qiao]{wu2022trajectory}
Penghao Wu, Xiaosong Jia, Li Chen, Junchi Yan, Hongyang Li, and Yu Qiao.
\newblock Trajectory-guided control prediction for end-to-end autonomous
  driving: A simple yet strong baseline.
\newblock \emph{Advances in Neural Information Processing Systems},
  35:\penalty0 6119--6132, 2022.

\bibitem[xAI(2024)]{grok}
xAI.
\newblock Grok-1.5 vision preview, 2024.

\bibitem[Xie et~al.(2023)Xie, Xu, Rakotosaona, Rim, Tombari, Keutzer, Tomizuka,
  and Zhan]{xie2023sparsefusion}
Yichen Xie, Chenfeng Xu, Marie-Julie Rakotosaona, Patrick Rim, Federico
  Tombari, Kurt Keutzer, Masayoshi Tomizuka, and Wei Zhan.
\newblock Sparsefusion: Fusing multi-modal sparse representations for
  multi-sensor 3d object detection.
\newblock In \emph{Proceedings of the IEEE/CVF International Conference on
  Computer Vision}, pages 17591--17602, 2023.

\bibitem[Xu et~al.(2024)Xu, Zhang, Xie, Zhao, Guo, Wong, Li, and
  Zhao]{xu2024drivegpt4}
Zhenhua Xu, Yujia Zhang, Enze Xie, Zhen Zhao, Yong Guo, Kwan-Yee~K Wong,
  Zhenguo Li, and Hengshuang Zhao.
\newblock Drivegpt4: Interpretable end-to-end autonomous driving via large
  language model.
\newblock \emph{IEEE Robotics and Automation Letters}, 2024.

\bibitem[Zhai et~al.(2023{\natexlab{a}})Zhai, Feng, Du, Mao, Liu, Tan, Zhang,
  Ye, and Wang]{zhai2023ADMLP}
Jiang-Tian Zhai, Ze Feng, Jihao Du, Yongqiang Mao, Jiang-Jiang Liu, Zichang
  Tan, Yifu Zhang, Xiaoqing Ye, and Jingdong Wang.
\newblock Rethinking the open-loop evaluation of end-to-end autonomous driving
  in nuscenes.
\newblock \emph{arXiv preprint arXiv:2305.10430}, 2023{\natexlab{a}}.

\bibitem[Zhai et~al.(2023{\natexlab{b}})Zhai, Mustafa, Kolesnikov, and
  Beyer]{zhai2023sigmoid}
Xiaohua Zhai, Basil Mustafa, Alexander Kolesnikov, and Lucas Beyer.
\newblock Sigmoid loss for language image pre-training.
\newblock In \emph{Proceedings of the IEEE/CVF International Conference on
  Computer Vision}, pages 11975--11986, 2023{\natexlab{b}}.

\bibitem[Zheng et~al.(2023{\natexlab{a}})Zheng, Chiang, Sheng, Zhuang, Wu,
  Zhuang, Lin, Li, Li, Xing, Zhang, Gonzalez, and Stoica]{zheng2023judging}
Lianmin Zheng, Wei-Lin Chiang, Ying Sheng, Siyuan Zhuang, Zhanghao Wu, Yonghao
  Zhuang, Zi Lin, Zhuohan Li, Dacheng Li, Eric.~P Xing, Hao Zhang, Joseph~E.
  Gonzalez, and Ion Stoica.
\newblock Judging llm-as-a-judge with mt-bench and chatbot arena,
  2023{\natexlab{a}}.

\bibitem[Zheng et~al.(2023{\natexlab{b}})Zheng, Chen, Huang, Zhang, Duan, and
  Lu]{zheng2023occworld}
Wenzhao Zheng, Weiliang Chen, Yuanhui Huang, Borui Zhang, Yueqi Duan, and Jiwen
  Lu.
\newblock Occworld: Learning a 3d occupancy world model for autonomous driving.
\newblock \emph{arXiv preprint arXiv:2311.16038}, 2023{\natexlab{b}}.

\bibitem[Zheng et~al.(2024)Zheng, Song, Guo, and Chen]{zheng2024genad}
Wenzhao Zheng, Ruiqi Song, Xianda Guo, and Long Chen.
\newblock Genad: Generative end-to-end autonomous driving.
\newblock \emph{arXiv preprint arXiv:2402.11502}, 2024.

\end{thebibliography}
}

\clearpage
\setcounter{page}{1}
\maketitlesupplementary

The supplementary material is organized as follows. Sec.~\ref{sec:implementation} provides implementation details of our proposed methods including the concrete input prompt and target output. 
Sec.~\ref{sec:behavior} provides the model performance for each ego-vehicle behavior separately. Sec.~\ref{sec:extra_qualitative} gives additional visualization of \ourmethod{} motion planning in diverse scenarios. Finally, Sec.~\ref{sec:extra_ablation} conducts extra ablation studies on \ourdataset{} benchmark to justify the design of \ourmethod{}.

\section{Implementation Details}
\label{sec:implementation}
\paragraph{\ourdataset{} benchmark.} This benchmark contains 487k scenarios for model training and 44k for validation, which are divided from 103k sequences of $20s$ length. Each scenario contains $1s$ history and $8s$ future. We only consider the future $5s$ for the open-loop evaluation of motion planning since motion planning in real task is conducted in an iterative manner. At each time stamp, the dataset provides multi-view images captured by 8 cameras (namely \textit{front left, front, front right, side left, side right, rear left, rear, rear right}). The dataset is also equipped with fine-grained labels for all the other agents and the roadgraph although they are not used in our self-supervised motion planning algorithm.

\paragraph{Ego-vehicle coordinate system.} We conduct the motion planning task in the ego-vehicle coordinate system. At current timestamp, the ego-vehicle position is defined as the origin. The $x-$axis is oriented towards the current heading angle of the ego-vehicle. The $y-$axis is oriented towards the left of the ego-vehicle. The $z-$axis is oriented upwards. Although it is a 3D coordinate system, we only consider the $xy-$plane for the description of the positions, velocities, and accelerations of the vehicles in the motion planning task.
\paragraph{Architecture and hyperparameters.} Unless otherwise specified, we build up the \ourmethod{} framework based on multimodal large-language model PaLI3-5B~\cite{chen2023pali}. The 3D scene representation (Sec.~\ref{sec:3dscene}) in default covers the range of $(-30m,80m)\times(-30m,30m)\times(-2m,8m)$ in the ego-vehicle coordinate along $x,y,z$ axes separately. The position embedding in Eq.~\ref{eq:projection} and Eq.~\ref{eq:sparse_project} are implemented with a two-layer MLP attached to the Fourier embedding of $(x,y,z)$. To generate the gate values, we reduce the dimension of visual features from $C=1536$ to $C'=96$ in Eq.~\ref{eq:reduce_dim}. The volume resolution is $1m\times 1m\times 2m$. For each scene, we select $M=6000$ sparse volumes based on the gate values. For attention bias (Sec.~\ref{sec:bias}), there are independent intra-modality biases for visual tokens and text tokens separately. We leave the details of bias function $b(\cdot)$ in the extra ablation study (Sec.~\ref{sec:extra_ablation}). For each attention head at each self-attention layer, we also include learnable scalar inter-modality biases for the attention values between text tokens and visual tokens. In Sec.~\ref{sec:temporal}, for temporal fusion, apart from the current frame, we also take camera images from $T=1$ historical frame at the timestamp $-0.5s$ as model input. For multi-decoding aggregation (Sec.~\ref{sec:multi}), the model outputs $K=16$ trajectories in parallel through Top-$p$ sampling (nucleus sampling)~\cite{holtzman2019curious} ($p=0.9$), which are aggregated as one unique planning result. For both datasets, the input images are resized to $448\times448$ as model inputs.

\paragraph{Model initialization.} To ensure the alignment within the pretrained MLLM, we apply the following special rules to relieve the disturbance of injected modules to the pretrained weights in the early finetuning stage. The last FC-layer of MLP for the position embeddings in Eq.~\ref{eq:projection} and Eq.~\ref{eq:sparse_project} are zero initialized. The learnable vacant feature $\mathbf{f}_{vac}$ for sparse volume representation is also initialized as zeros (Eq.~\ref{eq:vac}). We also set the initial bias for each bin in the bias function $b(\cdot)$ (Eq.~\ref{eq:bias}) as zeros (details in Sec.~\ref{sec:extra_ablation}). Finally, the temporal fusion (Sec.~\ref{sec:temporal}), we maintain the channel-wise semantics of visual features by initializing the weight matrix of the FC layer as $\mathbf{W}=\left[I\in\mathbb{R}^{C\times C};\mathbf{0}^{C\times (T-1)\cdot C}\right]$ and the bias as zero.  Based on the above rules, the inserted modules would not significantly change the channel-wise semantics of the visual features from original pretrained perspective view features, which allows for a stable and efficient finetuning process.

\begin{figure}[t!]
    \centering
    \includegraphics[width=\linewidth]{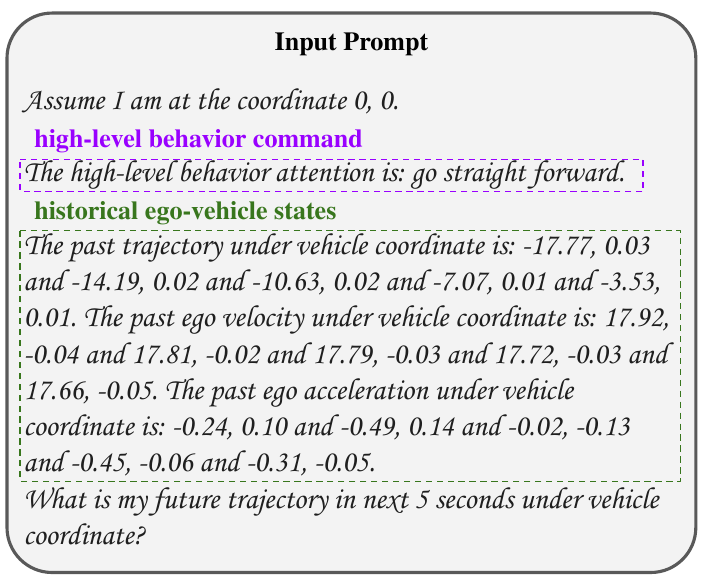}
    \includegraphics[width=\linewidth]{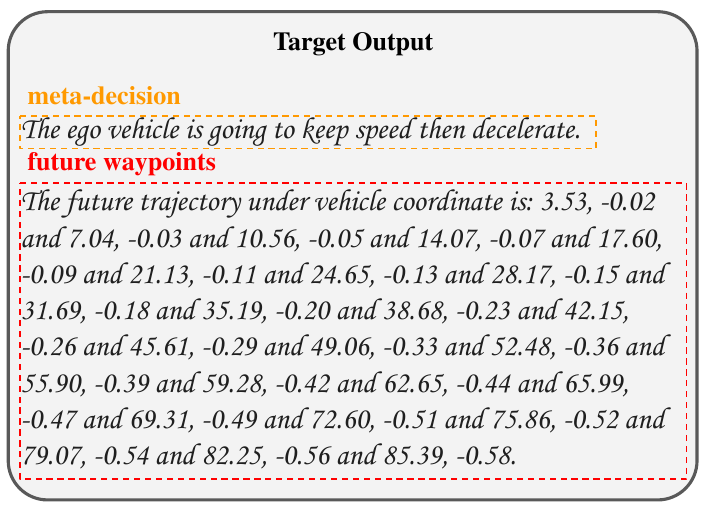}
    \caption{Example input prompt and target output on \ourdataset{}.}
    \label{fig:prompt}
\end{figure}

\paragraph{Input prompt and target output.}
The input prompt is composed of the high-level behavior command and ego-vehicle historical states. We represent all historical states in the language space. The position, velocity, and acceleration at each time stamp are represented with two floating numbers (two decimal places) separately for the $x-$component and $y-$component in the ego-vehicle coordinate. The target output is composed of meta-decision (Sec.~\ref{sec:decision} only in \ourmethod{}) and future waypoints. The future waypoints are also represented in the language space with two floating numbers for each time stamp. The historical and future states are sampled at a frequency of $5Hz$ on \ourdataset{} where we consider the historical states of past $1s$ and future waypoints of future $5s$. On nuScenes dataset, we consider the historical states of past $1s$ and future waypoints of future $3s$ at a frequency of $2Hz$. An example of the input and output on \ourdataset{} is visualized in Fig.~\ref{fig:prompt}.

\paragraph{Heuristics for high-level behaviors.} We consider the seven behaviors for the ego-vehicle on \ourdataset{} benchmark, which are determined based on the ego-vehicle ground-truth future trajectories following heuristic rules. 
\begin{enumerate}
    \item \textbf{Stop}: The ego-vehicle movement is $<5m$ and the maximal speed is $<2m/s$.
    \item \textbf{Do left turn}: The ego-vehicle does not stop. The final heading angle is $>30^{\circ}$. The final position is $\geq-5m$ along $x-$axis.
    \item \textbf{Do left U-turn}: The ego-vehicle does not stop. The final heading angle is $>30^{\circ}$. The final displacement is $<-5m$ along $x-$axis.
    \item \textbf{Do right turn}: The ego-vehicle does not stop. The final heading angle is $<-30^{\circ}$.
    \item \textbf{Go straight left}: The ego-vehicle does not stop. The final heading angle is in the range $[-30^{\circ}, 30^{\circ}]$. The final displacement is $>5m$ along $y-$axis.
    \item \textbf{Go straight right}: The ego-vehicle does not stop. The final heading angle is in the range $[-30^{\circ}, 30^{\circ}]$. The final displacement is $<-5m$ along $y-$axis.
    \item \textbf{Go straight forward}: The ego-vehicle does not stop. The final heading angle is in the range $[-30^{\circ}, 30^{\circ}]$. The final displacement along $y-$axis is in the range $[-5m,5m]$.
\end{enumerate}
For the calculation of \textit{bADE} metric in the model evaluation, we follow the above rules to determine the behavior based on the ground-truth future trajectories of $8s$ although our planning horizon is only $5s$. For the high-level behavior command in the model inputs, ``stop" is excluded to avoid future information leakage. To determine the high-level behavior command input, we start from the ground-truth $8s$ future trajectories. If none of behaviors 2-7 is satisfied, we would prolong the future horizon by $2s$ until at least one of behaviors 2-7 is satisfied. If the end of the collected trajectory sequence is reached, we would just consider that scenario as ``go straight forward".

\paragraph{Heuristics for meta-decisions.} In Sec.~\ref{sec:decision}, we design a meta-decision strategy as a preliminary prediction before the waypoints. The meta-decision of ego-vehicle includes four categories determined with following heuristic rules based on the ground-truth future information.
\begin{enumerate}
    \item \textbf{Keep stationary}: The maximal speed is $<2m/s$ and the final displacement is $<1.5m$.
    \item \textbf{Keep speed}: The ego-vehicle does not keep stationary. The average acceleration is in the range of $[-0.5m/s^2,0.5m/s^2]$.
    \item \textbf{Accelerate}: The ego-vehicle does not keep stationary. The average acceleration is $>0.5m/s^2$.
    \item \textbf{Decelerate}: The ego-vehicle does not keep stationary. The average acceleration is $<-0.5m/s^2$.
\end{enumerate}
The model should firstly predict the meta-decision and then auto-regressively output the future waypoints. On \ourdataset{} with $5s$ future prediction horizon, we divide the $5s$ into two stages of $2.5s$, where meta-decisions are predicted independently for each stage. On nuScenes dataset with $3s$ prediction horizon, we only consider one-stage meta-decision.

\begin{table*}[ht!]
    \centering
    \resizebox{\linewidth}{!}{
    \begin{tabular}{c|ccccccc|cc}
        \toprule
       \multirow{2}{*}{\textbf{Methods}}  & \multicolumn{7}{c|}{\textbf{ADE@5s for each behavior}} & \multirow{2}{*}{ADE@5s} & \multirow{2}{*}{bADE@5s}  \\
        & stop & straight forward & straight left & straight right & left turn & right turn & left U-turn & & \\
        \midrule
        Vanilla PaLI & \textbf{0.048} & 0.960 & 1.252 & 1.297 & 1.566 & 1.323 & 1.039& 0.798 & 1.069\\
        \ourmethod{} (ours) & 0.063 & 0.843 & 1.077 &1.177 &1.252 & 1.158& 0.925 & 0.693 & 0.928 \\ 
        \ourmethod{}* (ours) & 0.065 & \textbf{0.806} & \textbf{0.957} & \textbf{1.074} & \textbf{1.124} & \textbf{1.027} & \textbf{0.756} &\textbf{0.655} & 0\textbf{.830}\\
        \midrule
        \multicolumn{10}{c}{\textit{\textcolor{gray}{
        Models that take high-quality objects, tracks, and roadgraphs as inputs instead of using raw camera images.
        }}}\\
        \midrule
        \textcolor{gray}{MotionLM} & \textcolor{gray}{0.048}& \textcolor{gray}{0.832} & \textcolor{gray}{1.293} & \textcolor{gray}{1.275} & \textcolor{gray}{1.239} & \textcolor{gray}{1.172} & \textcolor{gray}{0.990} & \textcolor{gray}{0.697} &  \textcolor{gray}{0.978}\\
        \bottomrule
    \end{tabular}
    }
    \caption{Behavior-wise sliced metrics on \ourdataset{} benchmark. ``*" denotes methods with internal data pretraining.}
    \label{tab:sliced}
\end{table*}

\section{Behavior-wise Model Performance}
\label{sec:behavior}
In Tab.~\ref{tab:sliced}, we report the \textit{ADE@5s} metric of \ourmethod{} for each ego-vehicle behavior on \ourdataset{} benchmark separately. Results show the superiority of \ourmethod{} especially in complicated scenarios like turnings, where a significant performance gain is witnessed from vanilla PaLI baseline to \ourmethod{}. Motion planning for these difficult behaviors requires a great understanding of the roadgraph and other agents  from raw camera images, which reflects the strong reasoning ability of our proposed spatio-temporal visual representation.

\begin{figure*}
    \centering
    \begin{subfigure}{0.48\linewidth}
        \centering
        \includegraphics[width=0.32\linewidth]{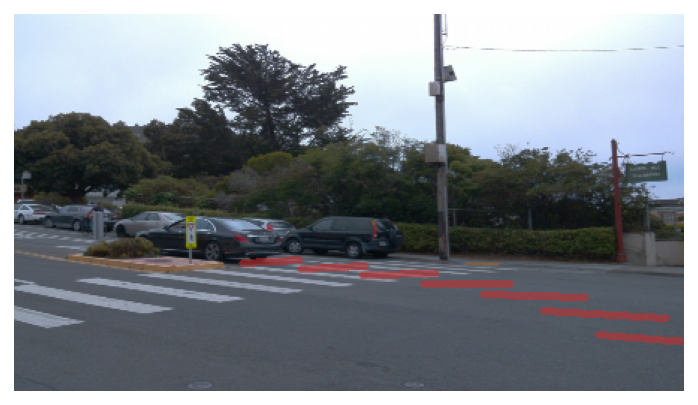}
        \includegraphics[width=0.32\linewidth]{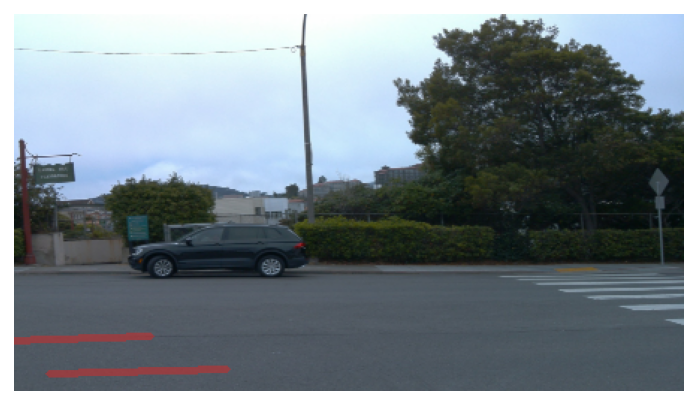}
        \includegraphics[width=0.32\linewidth]{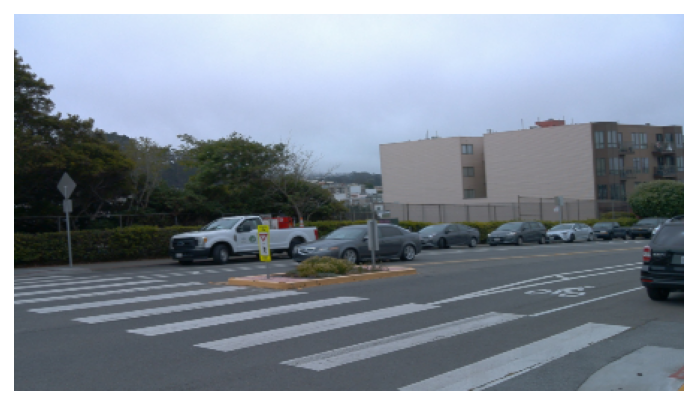}
    \end{subfigure}
    \hfill
    \begin{subfigure}{0.48\linewidth}
        \centering
        \includegraphics[width=0.32\linewidth]{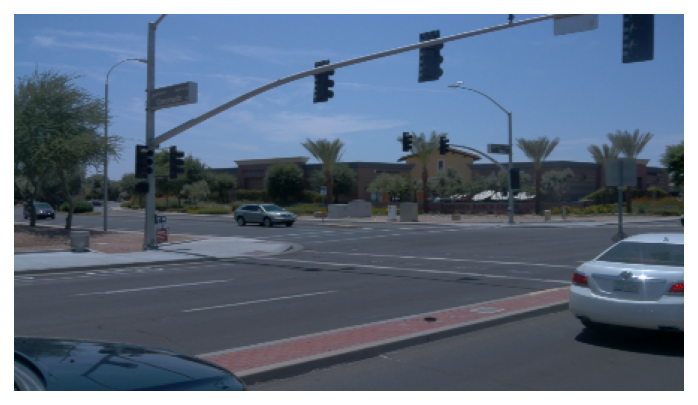}
        \includegraphics[width=0.32\linewidth]{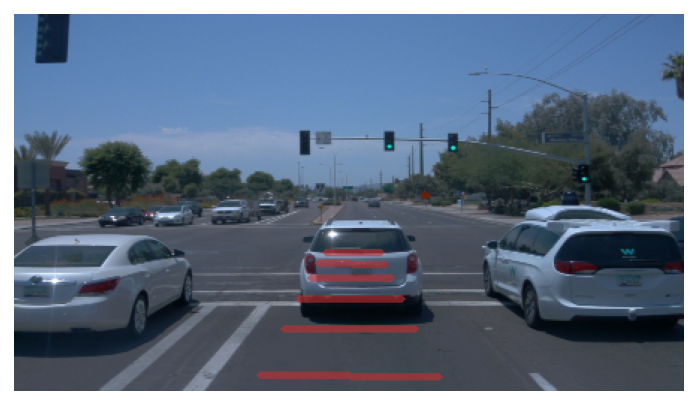}
        \includegraphics[width=0.32\linewidth]{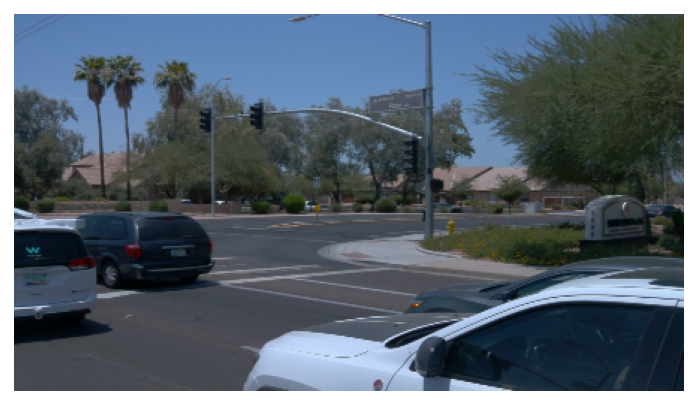}
    \end{subfigure}
    \begin{subfigure}{0.48\linewidth}
        \centering
        \includegraphics[width=0.32\linewidth]{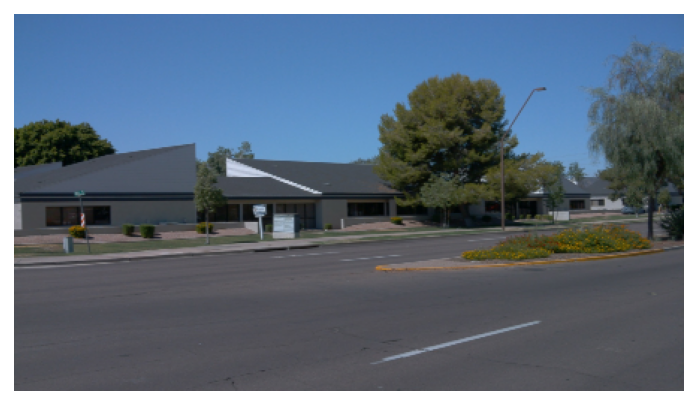}
        \includegraphics[width=0.32\linewidth]{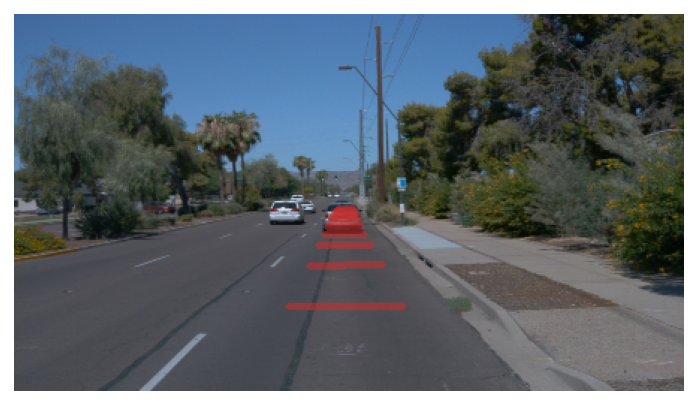}
        \includegraphics[width=0.32\linewidth]{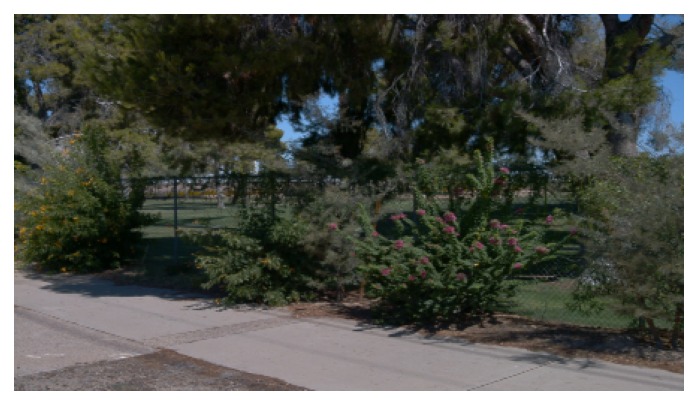}
    \end{subfigure}
    \hfill
    \begin{subfigure}{0.48\linewidth}
        \centering
        \includegraphics[width=0.32\linewidth]{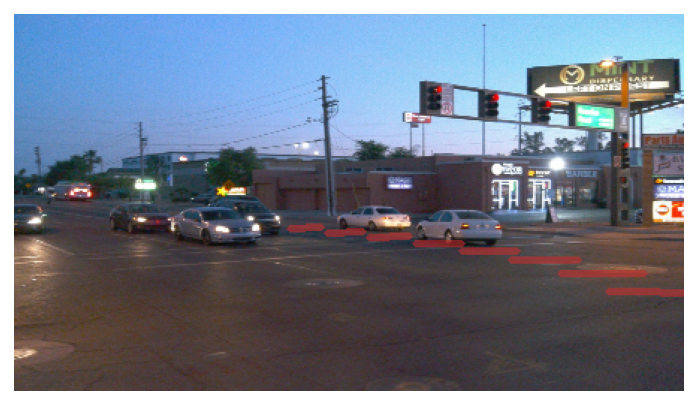}
        \includegraphics[width=0.32\linewidth]{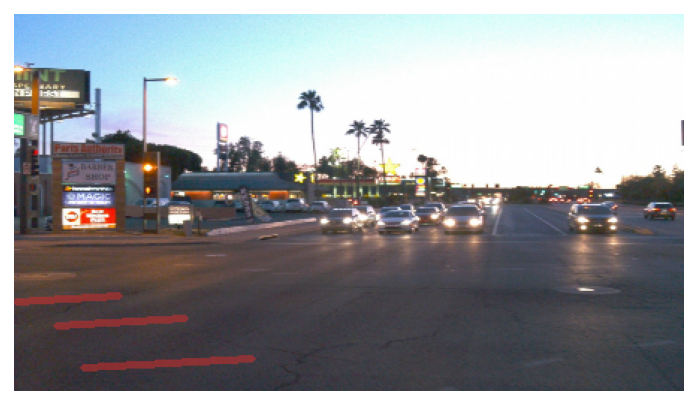}
        \includegraphics[width=0.32\linewidth]{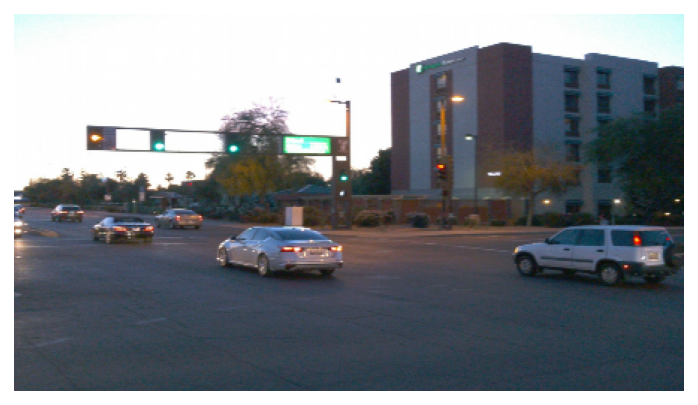}
    \end{subfigure}
    \begin{subfigure}{0.48\linewidth}
        \centering
        \includegraphics[width=0.32\linewidth]{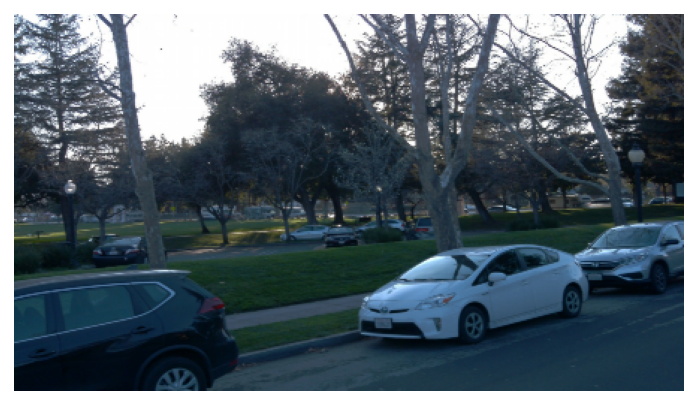}
        \includegraphics[width=0.32\linewidth]{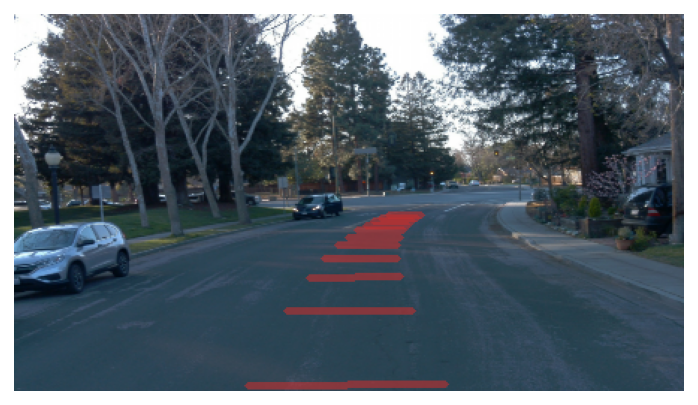}
        \includegraphics[width=0.32\linewidth]{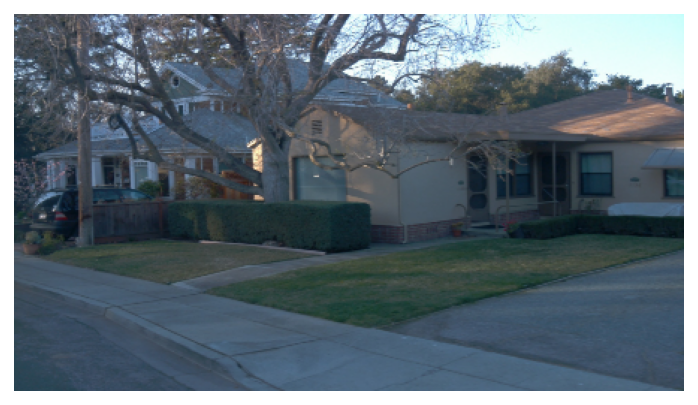}
    \end{subfigure}
    \hfill
    \begin{subfigure}{0.48\linewidth}
        \centering
        \includegraphics[width=0.32\linewidth]{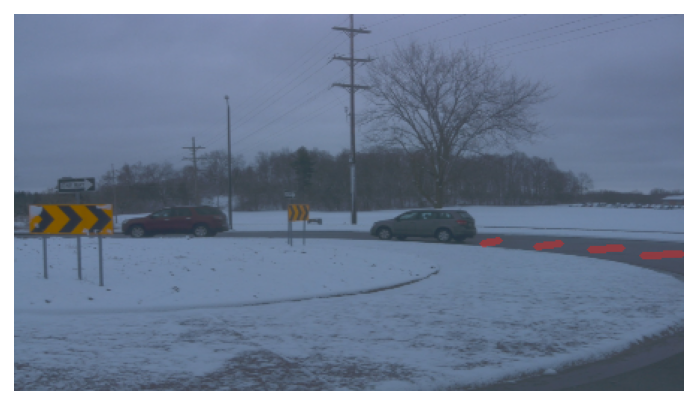}
        \includegraphics[width=0.32\linewidth]{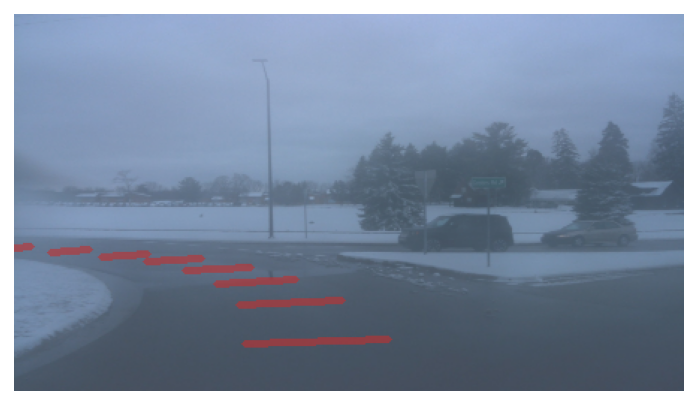}
        \includegraphics[width=0.32\linewidth]{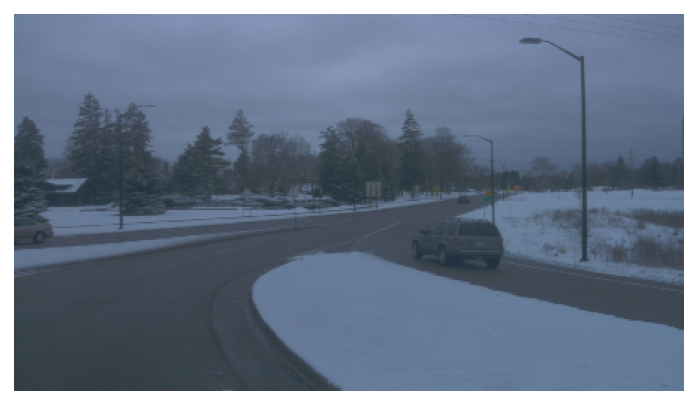}
    \end{subfigure}
    \begin{subfigure}{0.48\linewidth}
        \centering
        \includegraphics[width=0.32\linewidth]{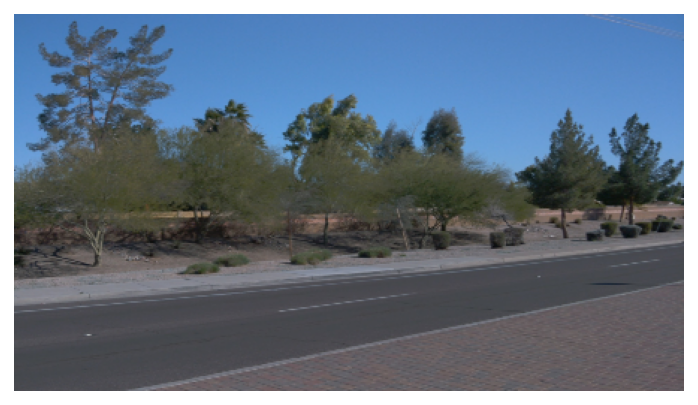}
        \includegraphics[width=0.32\linewidth]{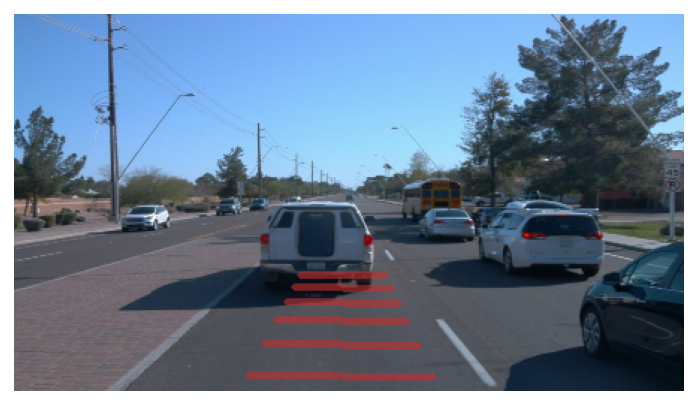}
        \includegraphics[width=0.32\linewidth]{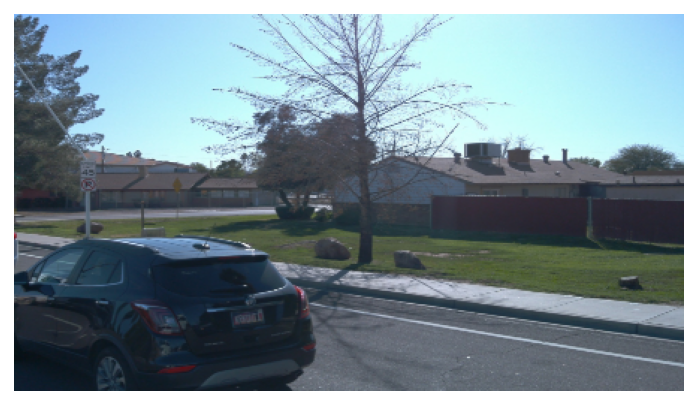}
    \end{subfigure}
    \hfill
    \begin{subfigure}{0.48\linewidth}
        \centering
        \includegraphics[width=0.32\linewidth]{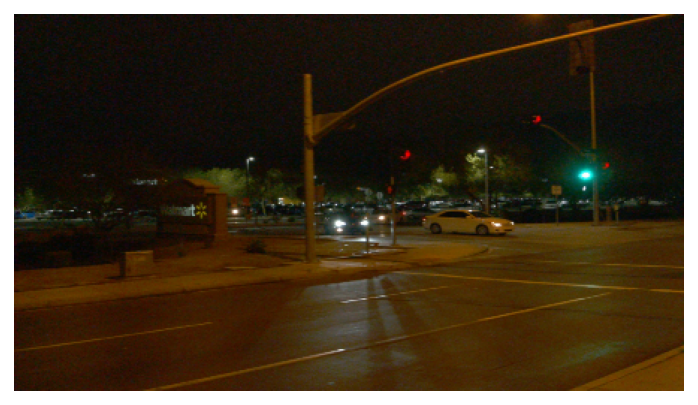}
        \includegraphics[width=0.32\linewidth]{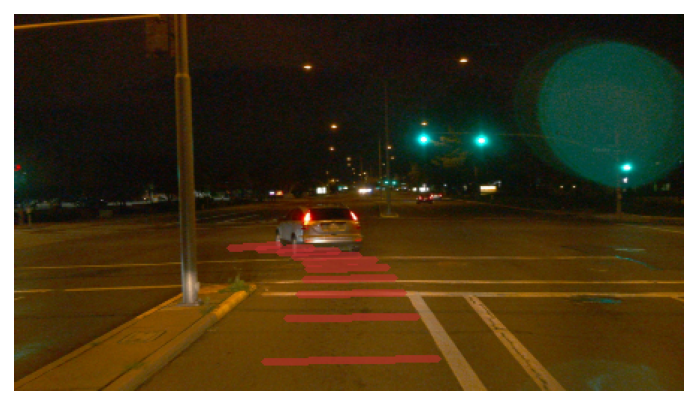}
        \includegraphics[width=0.32\linewidth]{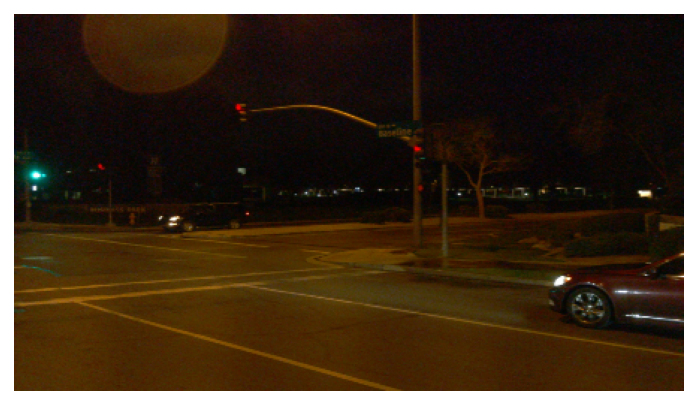}
    \end{subfigure}
    \begin{subfigure}{0.48\linewidth}
        \centering
        \includegraphics[width=0.32\linewidth]{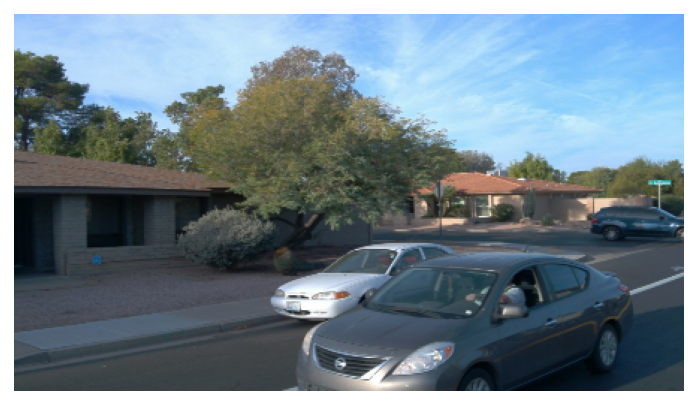}
        \includegraphics[width=0.32\linewidth]{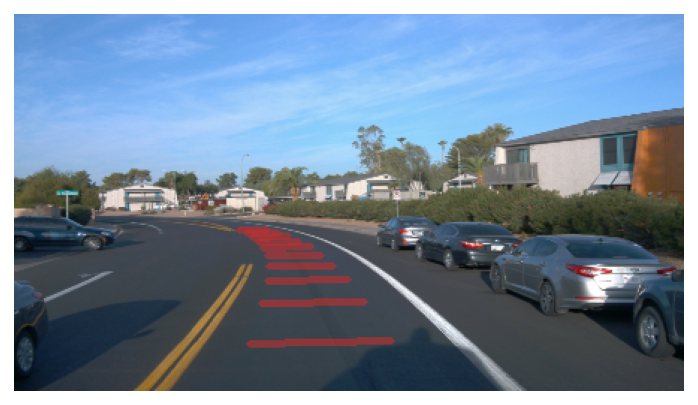}
        \includegraphics[width=0.32\linewidth]{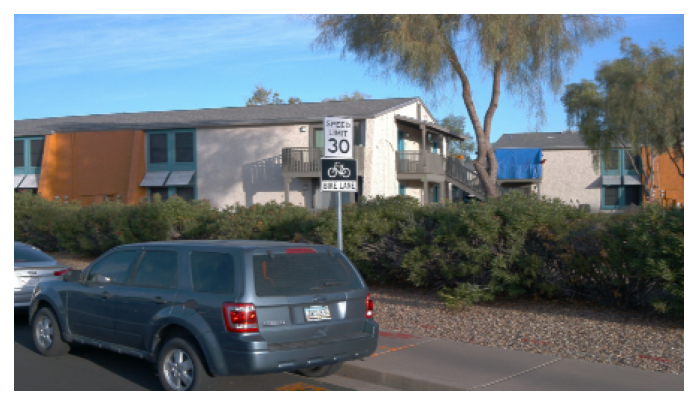}
    \end{subfigure}
    \hfill
    \begin{subfigure}{0.48\linewidth}
        \centering
        \includegraphics[width=0.32\linewidth]{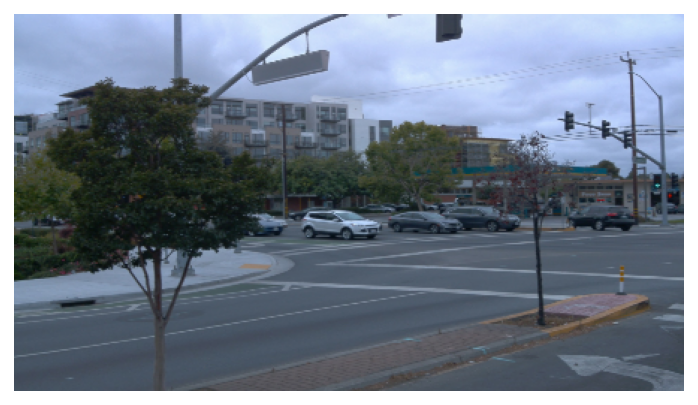}
        \includegraphics[width=0.32\linewidth]{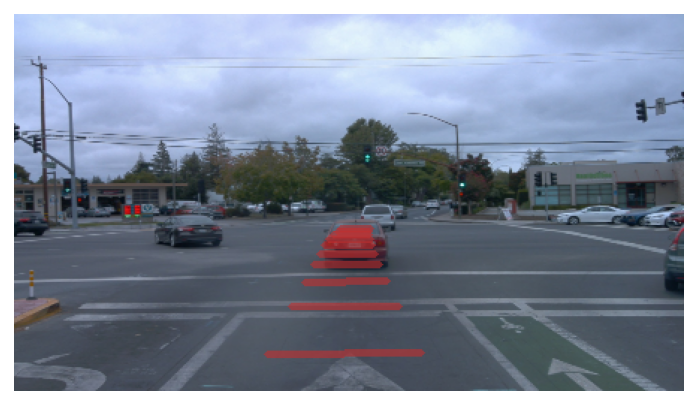}
        \includegraphics[width=0.32\linewidth]{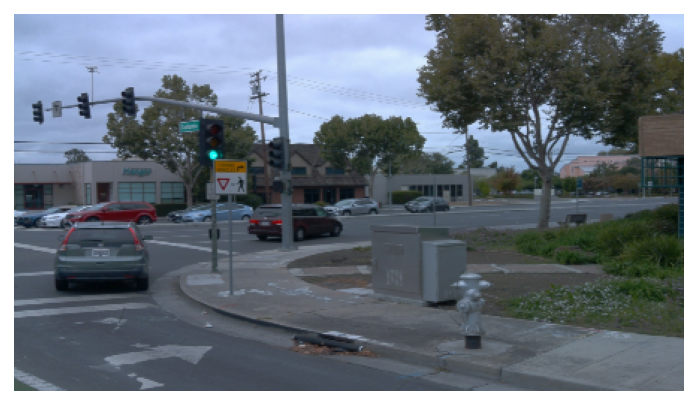}
    \end{subfigure}
    \begin{subfigure}{0.48\linewidth}
        \centering
        \includegraphics[width=0.32\linewidth]{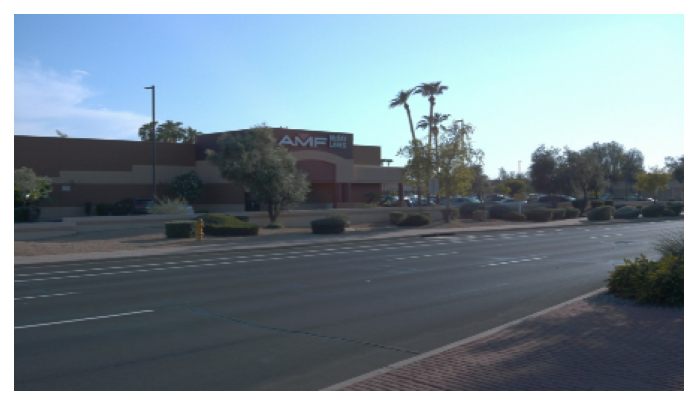}
        \includegraphics[width=0.32\linewidth]{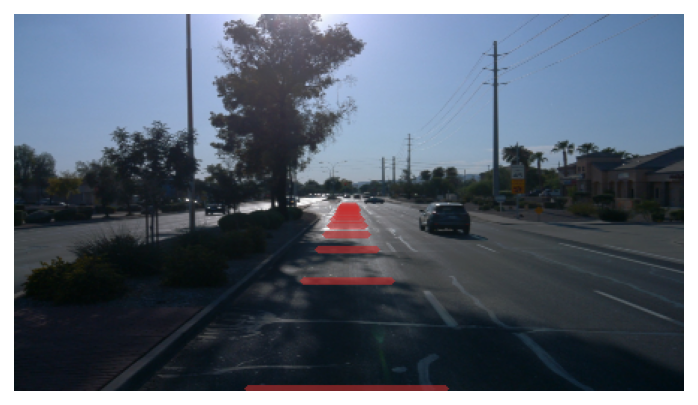}
        \includegraphics[width=0.32\linewidth]{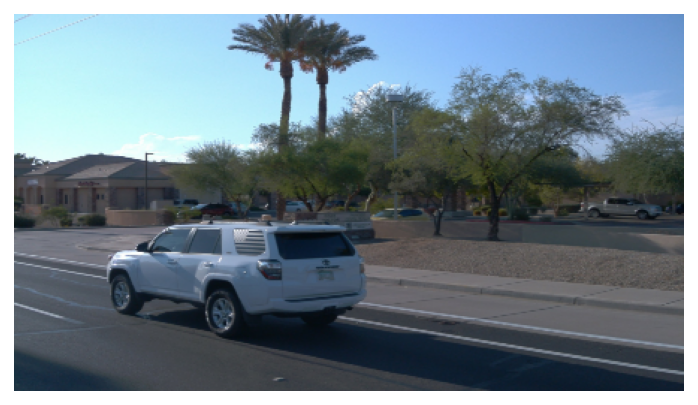}
    \end{subfigure}
    \hfill
    \begin{subfigure}{0.48\linewidth}
        \centering
        \includegraphics[width=0.32\linewidth]{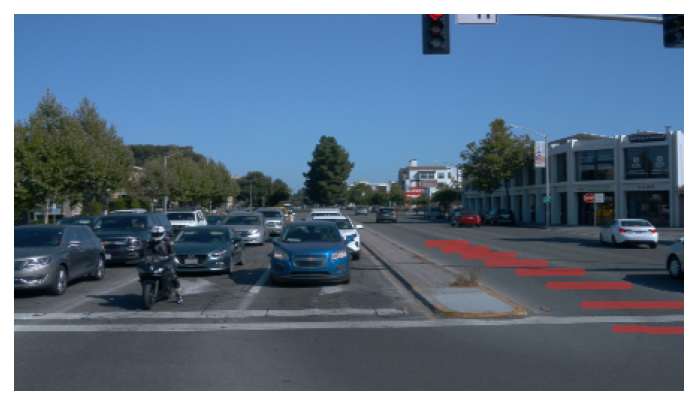}
        \includegraphics[width=0.32\linewidth]{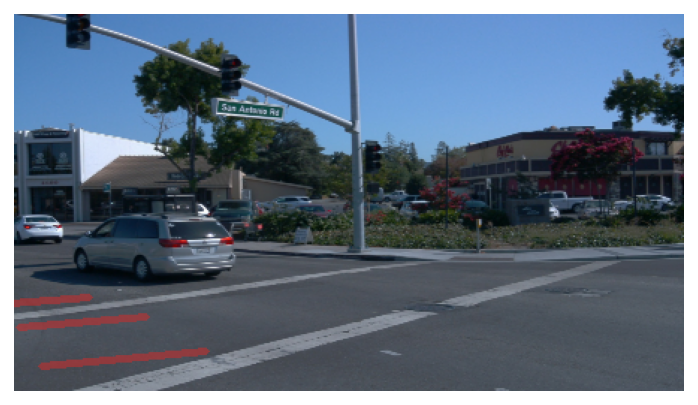}
        \includegraphics[width=0.32\linewidth]{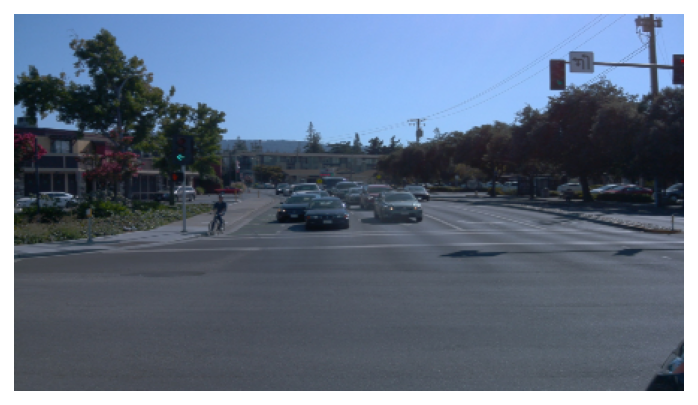}
    \end{subfigure}
    \begin{subfigure}{0.48\linewidth}
        \centering
        \includegraphics[width=0.32\linewidth]{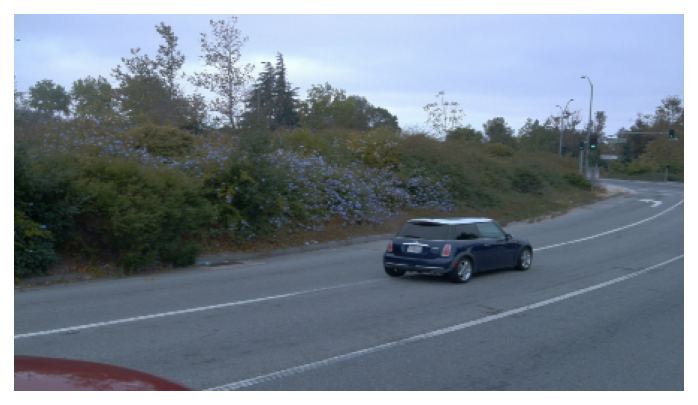}
        \includegraphics[width=0.32\linewidth]{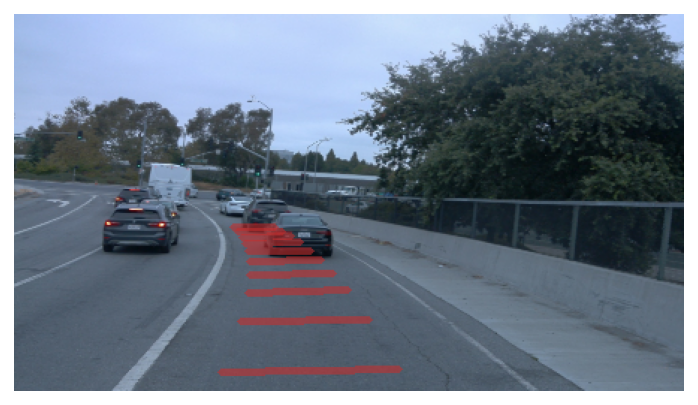}
        \includegraphics[width=0.32\linewidth]{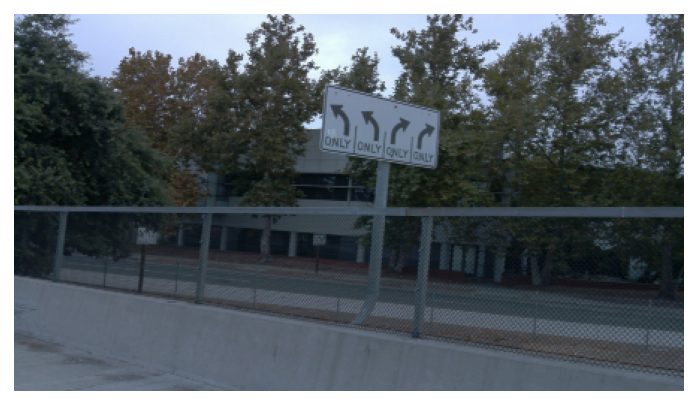}
    \end{subfigure}
    \hfill
    \begin{subfigure}{0.48\linewidth}
        \centering
        \includegraphics[width=0.32\linewidth]{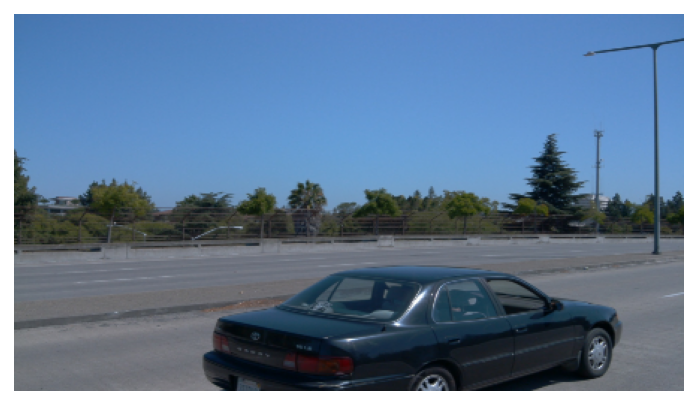}
        \includegraphics[width=0.32\linewidth]{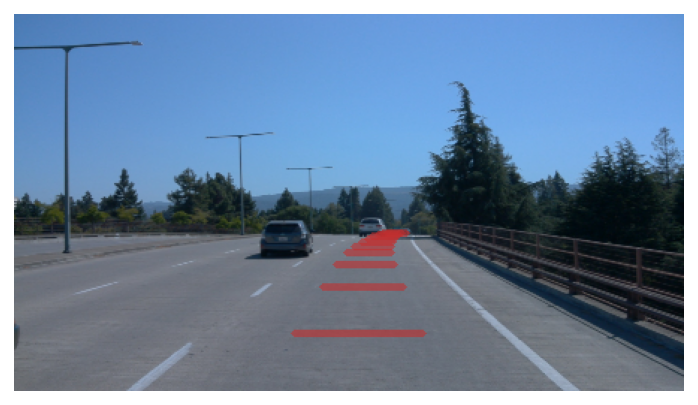}
        \includegraphics[width=0.32\linewidth]{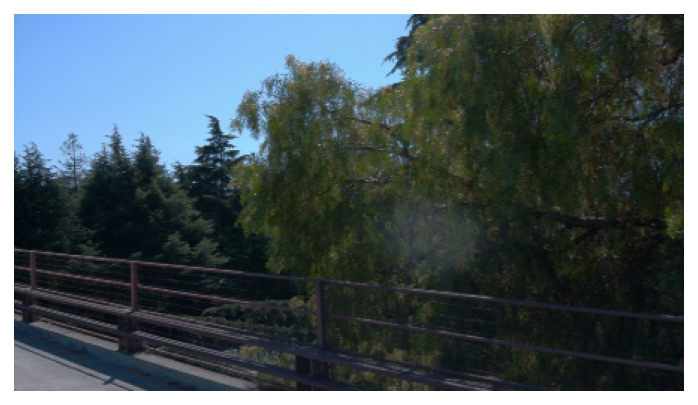}
    \end{subfigure}
    \caption{Additional qualitative results of motion planning. We show the \textit{front left, front, front right} cameras for each case.}
    \label{fig:extra_vis}
\end{figure*}

\section{Additional Qualitative Results}
\label{sec:extra_qualitative}
In Fig.~\ref{fig:extra_vis}, we visualize more planning results on \ourdataset{}. Examples cover different behaviors, speeds, lighting conditions, and weathers. Results show the robust performance of \ourmethod{} in all these diverse scenarios.

\section{Additional Ablation Studies}
\label{sec:extra_ablation}
In this part, we conduct several additional ablation studies to further justify the design of our \ourmethod{} including the camera configuration, relative attention bias, multi-decoding aggregation, and motion tokens. 

\noindent\textbf{Camera configuration.}
We apply different configurations of camera sensors in Tab.~\ref{tab:ablation-camera}. Consistent with intuitions, the \textit{front} camera is the most important, which can solely guarantee a reasonable planning performance. Adding other cameras continuously improves the model performance since they provide more and more complete information of the surrounding environment. The three cameras in the back can notably boost the planning model since they provide cues about other agents behind ego-vehicle. This information helps to determine the future ego-vehicle velocities and accelerations.

\begin{table}[t]
    \centering
    \resizebox{\linewidth}{!}{
    \begin{tabular}{c|cc}
       \toprule
        \textbf{Camera configuration} & \textbf{ADE@5s} & \textbf{bADE@5s}\\
       \midrule
        \textit{front} & 0.765 & 1.036   \\
        \textit{front, front left/right} & 0.751 & 1.012 \\
        \textit{front, front left/right, side left/right} & 0.746 & \textbf{0.981} \\
        \midrule
        \textit{all eight surrounding cameras} & \textbf{0.732} & 0.985\\
    \bottomrule
    \end{tabular}
    }
    \vspace{-10pt}
    \caption{Ablation studies on camera configurations.}
    \label{tab:ablation-camera}
\end{table}

\paragraph{Relative attention bias.} We consider two types of relative position bias function $b(\cdot)$ in Eq.~\ref{eq:bias}.
\begin{enumerate}
    \item \textbf{Linear bias:} The bias is linearly related to the relative distance between tokens. For the sparse volume visual tokens, $b(x,y,z)=b_x(\Delta x)+b_y(\Delta y)+b_z(\Delta z)$, we take the $x-$axis as an example,
    \begin{equation}
        b_x(\Delta x)=\tau_x\cdot|\Delta x|,
    \end{equation}
    where $\tau_x$ is a learnable parameter separately for each attention head at each layer, and $\Delta x$ is the relative position between sparse volume tokens along $x-$axis.
    Similarly, for 1D text tokens, the separate bias function is 
    \begin{equation}
        b_p(\Delta p)=\tau_p\cdot|\Delta p|
    \end{equation}
    where $\tau_p$ is a learnable parameter separately for each attention head in each layer, and $\Delta p$ is the relative position between text tokens.
    \item \textbf{Bin-wise bias (our choice):} The relative positions $\Delta x, \Delta y, \Delta z$ (visual tokens) and $\Delta p$ (text tokens) for each axis are divided into 32 bins independently. For each axis, 16 bins cover the range $[-8(m),8(m)]$ linearly with an interval $1(m)$\footnote{The unit (m) is only for visual tokens ($\Delta x, \Delta y, \Delta z$) throughout this paragraph.}. The other 16 bins symmetrically cover the range $(-128(m),-8(m))$ and $(8(m),128(m))$ in log scale, where the relative distances are truncated to at most $128(m)$. In this case, if we take $b_x(\cdot)$ as an example, the relative bias function is written as 
    \begin{equation}
        b_x(\Delta x)=m_x(\text{bin}(\Delta x))
    \end{equation}
    where $m_x(\cdot)$ maps each bin to a learnable bias value independently for each attention head at each self-attention layer.
\end{enumerate}
Tab.~\ref{tab:ablation-bias} justifies the design of bin-wise bias, which brings greatly better performance. We think it is important to distinguish the two directions along each axis (\textit{e.g.} front or back) in the motion planning task. Besides, local feature aggregation is more sensitive to close neighbors at each locality, so the bin-wise bias function has finer grains in close distance compared to the linear bias function.

\begin{table}[ht]
    \centering
    \begin{tabular}{c|cc}
    \toprule
        \textbf{Attention bias} & \textbf{ADE@5s} & \textbf{bADE@5s} \\
    \midrule
        no bias & 0.750 & 1.005 \\
        linear bias function & 0.770 & 1.082\\
        bin-wise bias function & \textbf{0.732} & \textbf{0.985}\\
    \bottomrule
    \end{tabular}
    \vspace{-10pt}
    \caption{Ablation studies on attention bias.}
    \label{tab:ablation-bias}
    \vspace{-10pt}

\end{table}

\paragraph{Multi-decoding aggregation.}
We dig into the multi-decoding strategy which can bring notable performance gain. The motivation is that the MLLM is prone to assigning high confidence scores to simple future behaviors such as \textit{stop}. To this end, we encourage the model to output multiple future trajectories. Their aggregation serves as the final planning result, which can counteract the model's high confidence in simple behaviors. In Tab.~\ref{tab:ablation-multi}, we find that 1) The aggregation of more decoded trajectories leads to better performance. 2) Nucleus sampling can outperform beam search since it can generate more diverse outputs. 3) Weighted average is inferior to mean average since the model is prone to assigning high confidence to simple degenerated behaviors.

\begin{table}[ht!]
    \centering
    \resizebox{\linewidth}{!}{
    \begin{tabular}{ccc|cc}
    \toprule
        \textbf{Sample number} & \textbf{Sample strategy} & \textbf{Aggregation} & \textbf{ADE@5s} & \textbf{bADE@5s} \\
    \midrule
        1 & greedy sampling & - & 0.728 & 0.986\\
        4 & beam search & average & 0.739 & 0.997\\
        4 & nucleus sampling & average & 0.709 & 0.941 \\
        4 & nucleus sampling & weighted average & 0.747& 1.005\\
        16 & nucleus sampling & average & \textbf{0.693}& \textbf{0.928}\\
    \bottomrule
    \end{tabular}
    }
    \vspace{-10pt}
    \caption{Ablation studies on multi-decoding aggregation.}
    \label{tab:ablation-multi}
    \vspace{-10pt}

\end{table}

\begin{table}[ht!]
    \centering
    \begin{tabular}{c|cc}
    \toprule
        \textbf{Trajectory representation} & \textbf{ADE@5s} & \textbf{bADE@5s} \\
    \midrule
        motion tokens & 0.779& 1.061\\
        floating numbers & \textbf{0.750} & \textbf{1.005} \\
     \bottomrule
    \end{tabular}
    \vspace{-10pt}
    \caption{Ablation studies on motion trajectory tokenization.}
    \label{tab:ablation-token}
    \vspace{-10pt}

\end{table}

\paragraph{Motion tokens.}
Several prior works~\cite{seff2023motionlm,simadrivelm} benefit from specialized trajectory tokenization modules, which converts motion trajectories into extra discrete tokens added to the vocabulary of language models. However, in Tab.~\ref{tab:ablation-token}, the trajectory tokenization strategy similar with MotionLM~\cite{seff2023motionlm} hurts the performance of our \ourmethod{} in comparison with na\"{\i}ve floating number waypoints representation (Fig.~\ref{fig:prompt}). We also witness a much slower convergence speed with this extra trajectory tokenization. Since the MLLM is already pretrained on large-scale data, additionally injected trajectory tokens may not align with the pretrained model. In contrast, the floating number representation can align the historical and future states in the language space to exploit the large-scale MLLM pretraining with lower requirement for finetuning.

\end{document}